\documentclass[11pt,journal,comsoc, onecolumn]{IEEEtran}

\usepackage{amsmath, amsfonts, amssymb,  epsfig, mathtools, caption}
\usepackage{times, amsbsy,bm,amsthm}
\usepackage{framed} 
\usepackage{adjustbox}
\usepackage{algorithm}
\usepackage{algorithmic}

\usepackage{multirow, booktabs, hhline}
\usepackage{subcaption, xcolor}

\makeatletter
\newcommand{\removelatexerror}{\let\@latex@error\@gobble}
\makeatother
\DeclareMathOperator*{\argmin}{\arg\!\min}

\begin{document}
		
		\title{AdaER: An Adaptive Experience Replay Approach for Continual Lifelong Learning }

%
	\author{Xingyu~Li,~Bo~Tang,~\IEEEmembership{Senior Member,~IEEE},~and~Haifeng~Li,~\IEEEmembership{Member,~IEEE}\thanks{Copyright (c) 2015 IEEE. Personal use of this material is permitted. However, permission to use this material for any other purposes must be obtained from the IEEE by sending a request to pubs-permissions@ieee.org.}
		\thanks{Xingyu Li is with the Department of Electrical and Computer Engineering, Mississippi State University, Mississippi State, MS, 39762 USA. e-mail: xl292@msstate.edu.}%
  \thanks{Bo Tang is with the Department of Electrical and Computer Engineering,  Worcester Polytechnic Institute, Worcester, MA. 01609 USA, e-mail: btang1@wpi.edu}
		\thanks{Haifeng Li is with the School of Geosciences and fInfo-Physics, Central South University, Changsha 410083, China e-mail: lihaifeng@csu.edu.cn.}%
	}
	
  
\IEEEtitleabstractindextext{
		\begin{abstract}
		Continual lifelong learning is an machine learning framework inspired by human learning, where learners are trained to continuously acquire new knowledge in a sequential manner. However, the non-stationary nature of streaming training data poses a significant challenge known as catastrophic forgetting, which refers to the rapid forgetting of previously learned knowledge when new tasks are introduced. While some approaches, such as experience replay (ER), have been proposed to mitigate this issue, their performance remains limited, particularly in the class-incremental scenario which is considered natural and highly challenging. In this paper, we present a novel algorithm, called adaptive-experience replay (AdaER), to address the challenge of continual lifelong learning. AdaER consists of two stages: memory replay and memory update. In the memory replay stage, AdaER introduces a contextually-cued memory recall (C-CMR) strategy, which selectively replays memories that are most conflicting with the current input data in terms of both data and task. Additionally, AdaER incorporates an entropy-balanced reservoir sampling (E-BRS) strategy to enhance the performance of the memory buffer by maximizing information entropy.        To evaluate the effectiveness of AdaER, we conduct experiments on established supervised continual lifelong learning benchmarks, specifically focusing on class-incremental learning scenarios. The results demonstrate that AdaER outperforms existing continual lifelong learning baselines, highlighting its efficacy in mitigating catastrophic forgetting and improving learning performance.
		\end{abstract}
		
		}
	\maketitle
 \IEEEdisplaynontitleabstractindextext

\IEEEpeerreviewmaketitle


%

	\section{Introduction}
	
\IEEEPARstart{S}{tate}-of-the-art machine learning (ML) approaches have achieved remarkable performance in various tasks as image classification \cite{krizhevsky2012imagenet}, distributed optimization \cite{qu2022generalized, zhou2022deep}, and security \cite{li2021lomar}. However, when trained with new tasks from non-stationary distributions, these models tend to rapidly forget previously learned information, a phenomenon known as ``catastrophic forgetting" \cite{mccloskey1989catastrophic, goodfellow2013empirical}. In contrast, human brains possess the ability to learn different concepts and perform conflicting tasks in a lifelong sequential manner, which is a desirable characteristic for artificial intelligent systems. As a result, there has been a growing interest in the field of continual lifelong learning \cite{ kirkpatrick2017overcoming, nguyen2017variational}, aiming to train artificial learners with non-stationary streaming training data, temporally correlated inputs, and minimal supervision.
	
	One commonly used approach to address catastrophic forgetting is the utilization of previously trained experience through ``memory replay," which involves rehearsing the memory of previously learned tasks along with new incoming tasks to reactivate relevant knowledge in the learning model, promoting knowledge consolidation \cite{oudiette2013upgrading,van2016hippocampal}. Replay-based methods can be categorized into two groups based on how previous memories are used: experience replay, which stores raw training examples in a limited memory buffer \cite{chaudhry2019tiny}, and generative replay, which trains a separate generative model to generate synthetic samples for previously learned tasks \cite{shin2017continual}.
	
	Although there have been debates about the utilization of seen experiences in replay-based methods, recent studies suggest that these settings are necessary, especially in more challenging continual learning scenarios \cite{van2019three}. For instance, existing approaches struggle with the class-incremental (class-IL) scenario, where the learner needs to perform all learned tasks independently, as opposed to the simpler task-incremental (task-IL) problem where the learner makes decisions for a single task. The class-IL scenario, being more realistic and challenging, has gained popularity in recent years. Surprisingly, replay-based methods, such as experience replay (ER) \cite{rolnick2018experience}, outperform other approaches in the class-IL scenario. Generally, ER and ER-based methods involve two main stages: update and replay, which determine how unseen data is added to memory and which samples should be replayed, respectively. However, current ER methods have faced criticism for their random sampling strategies in both memory update and replay, and addressing this challenge remains an open problem in the field \cite{aljundi2019online}.
	
\begin{figure}[t!]
		\centering
		\includegraphics[width=0.9\columnwidth]{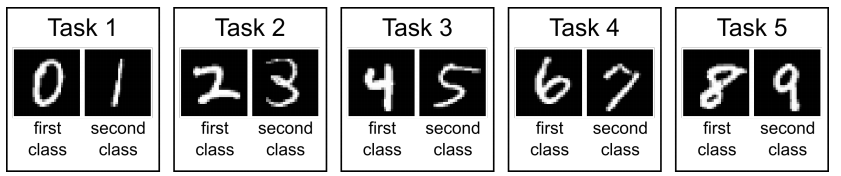}
		\caption{Split-MNIST: continual lifelong learning example.}
		\label{fig:class-IL}
	\end{figure}

	To overcome these limitations, we propose a novel continual learning algorithm called Adaptive Experience Replay (AdaER), which aims to enhance the efficiency of existing experience replay (ER) based approaches. In the replay stage, AdaER introduces the Contextually-Cued Memory Recall (C-CMR) method, which selects memories for replay based on contextual cues instead of random sampling. These contextual cues are derived from the most interfering examples (i.e., data-conflicting) and the associated forgetting tasks (i.e., task-conflicting) in the memory buffer. For instance, as illustrated in Fig.~\ref{fig:class-IL}, when the learner faces the 4th task of classifying digits 6 and 7, the performance of the previously learned classification of class 1 may be affected. In this case, replaying the memory of class 1 can help prevent forgetting. Similarly, the interference from one class can weaken the previously learned decision boundary within the task, making it beneficial to recall the memory of class 0 as well.
	
	Furthermore, AdaER improves the memory buffer updating strategy by maximizing the information entropy using the entropy-balanced reservoir sampling (E-BRS) method. In real-world scenarios, sequential streaming training data often exhibit imbalanced distributions, posing an additional challenge in continual learning. For example, the number of training examples for minority classes in the memory buffer may be limited, exacerbating the forgetting issue. The E-BRS method provides a balanced memory updating strategy that mitigates the bias caused by imbalanced training data. Through extensive experiments on multiple benchmarks, the results demonstrate that AdaER outperforms existing continual lifelong learning baselines. The contributions of this work can be summarized as follows:
	\begin{itemize}
		\item We propose the AdaER algorithm as a novel solution to address challenges in ER-based continual lifelong learning.
		\item For the replay stage of AdaER, we introduce the C-CMR method, which selects memories for replay based on contextual cues related to interference and task performance.
		\item The E-BRS method is developed to enhance the update stage of AdaER, improving the memory buffer's performance by maximizing information entropy.
	\end{itemize}

	The rest of this paper is organized as follows: the recent studies of continual lifelong learning are summarized in Sec.~\ref{Sec:related}. The problem statement and some preliminaries are given in Sec.~\ref{Sec:framework}. The proposed AdaER algorithm with C-CMR replay and E-BRS update methods is demonstrated in detail in Sec.~\ref{Sec:method}. Moreover, A comprehensive experiment study is presented in Sec.~\ref{Sec:analysis}, followed by a conclusion in Sec. VI.

	\section{Related Work}\label{Sec:related}

	\subsection{Continual Learning Approaches}
	\subsubsection{Memory Replay}
Replay-based methods draw inspiration from the relationship between the mammalian hippocampus and neocortex in neuroscience, aiming to replicate the interleaving of current training tasks and previously learned memories. This interplay between real episodic memories and generalized experiences offers valuable insights into knowledge consolidation. Early examples of this concept include the use of a dual-memory learning system in \cite{hinton1987using} to mitigate forgetting. More recent approaches, such as the modified self-organizing map (SOM) with short-term memory (STM) in \cite{gepperth2016incremental}, have been developed as part of incremental learning frameworks.

Taking inspiration from the generative role of the hippocampus, \cite{shin2017continual} proposed a dual-model architecture consisting of a generative model and a continual learning solver, enabling the sampling and interleaving of trained examples, known as Generative Replay (GR). Gradient Episodic Memory (GEM) \cite{lopez2017gradient} stores a subset of seen examples as episodic memory that has a positive impact on previous tasks, while Averaged GEM (A-GEM) \cite{chaudhry2019tiny} improves the computational and memory efficiency of GEM through an averaging mechanism. Experience Replay (ER) \cite{rolnick2018experience} uses reservoir sampling \cite{vitter1985random} to update the memory buffer, thereby approximating the data distribution. Works such as \cite{aljundi2019online} have further improved the memory update and replay processes to enhance performance, and FoCL \cite{lao2021focl} focuses on the feature space regularization instead of parameter space. \cite{zhuang2022multi} provides a  multi-criteria subset selection strategy to overcome the unstable problem of ER. 
Other ML topics as Meta-learning \cite{martins2023meta} and segmentation \cite{qiu2023sats} have also been applied to replay-based continual learning approaches. Despite being memory-intensive, replay-based methods have generally shown high performance.
	
	\subsubsection{Other Continual Learning Approaches}
	Apart from memory replay-based approaches, other categories of continual learning approaches include regularization methods and dynamic architecture methods. While our focus in this paper is on memory replay-based approaches, we briefly discuss these other categories below.

    Regularization methods aim to mitigate forgetting by retraining the lifelong learning model while balancing the knowledge of previous tasks and the current task. Learning without Forgetting (LwF) \cite{li2017learning} achieves this by distilling knowledge from a large model to a smaller model, ensuring that the predictions for the current task align with those of previously learned tasks. Elastic Weight Consolidation (EWC) \cite{kirkpatrick2017overcoming} identifies important weights for previously seen examples using Fisher Information and restricts their changes through quadratic terms in the loss function. Synaptic Intelligence (SI) \cite{zenke2017continual} penalizes parameters in the model's objective function in an unequal manner based on gradient information, identifying influential parameters. ISYANA \cite{mao2021continual} considers the relationship between tasks and the model, as well as the relationship between different concepts. \cite{sun2023exemplar} adopts variational auto-encoders to achieve exemplar-free continual lifelong learning.
    Regularization approaches are known for their ability to continually learn new tasks without storing seen examples or expanding the model's architecture. However, the trade-off in the loss function can lead to complex performance dynamics between seen and new tasks, especially when the task boundaries are unknown.
    
    Dynamic architecture approaches modify the model's architecture to accommodate new tasks by adding new neural resources. Some works, such as \cite{ yao2010boosting}, adopt a linear growth of the number of models in response to new tasks. Progressive Neural Networks (PNN) \cite{rusu2016progressive} preserves the previously trained network and allocates new sub-networks with fixed capacity to learn new information. Dynamically expanding network (DEN) \cite{yoon2017lifelong} incrementally increases the number of trainable parameters to adapt to new examples, providing an online method for expanding network capacity. Dynamic architecture approaches offer the advantage of preserving knowledge of seen tasks with fixed model parameters. However, they face challenges in preventing parameter growth from becoming too rapid, which could lead to complexity and resource demands. Additionally, selecting appropriate parameters to target the test task is a significant challenge in these approaches.

	\subsection{Three Scenarios of Continual Learning}
	The evaluation of continual lifelong learning approaches can be challenging due to differences in experimental protocols and access to task identity during testing. To facilitate meaningful comparisons, recent work by \cite{van2019three} has introduced three standardized evaluation scenarios of increasing difficulty for continual lifelong learning. These scenarios have been adopted by several subsequent studies \cite{buzzega2020dark}.

    We illustrate these scenarios using the popular continual learning benchmark, split MNIST \cite{zenke2017continual}, where the ten handwritten digits are learned sequentially in multiple tasks with limited classes, as shown in Figure \ref{fig:class-IL}. The first scenario is task-incremental learning (\textbf{task-IL}), which is the easiest scenario where the learner always knows the target learning task. The second scenario is \textbf{domain-IL}, where the task identity is partially known. For example, the learner needs to determine whether testing digits are even or odd in Figure \ref{fig:class-IL}. The most challenging scenario is \textbf{class-IL}, where the learner is required to distinguish all previously seen classes without knowing the task identity. Recent reports \cite{van2020brain} indicate that the class-IL scenario is more realistic and poses a greater challenge for incrementally learning new classes. In this paper, we specifically focus on enhancing the performance of memory replay-based continual learning approaches in the class-IL scenario.

	\section{Problem Statement}\label{Sec:framework}

	In traditional supervised learning, the goal of ML learners is to train a classifier $f$, parameterized by ${\theta}$ to address an optimization problem over a stationary dataset $\mathcal{X}$, which contains a number of training sample pairs $\{\bm{x}, y\}$ from a distribution $\mathcal{D}$
	that:
	\begin{equation}\label{Eq:stationary}
		{\theta}^{*}  = \argmin_{{\theta}} \mathbb{E}_{(\bm{x}, y) \sim \mathcal{D}} [l(f_\theta(\mathbf{x}), y)],
	\end{equation}
	where $l(\cdot, \cdot)$ is the loss function that denotes the empirical risk of $f_\theta$ over $\mathcal{D}$. Although these stationary optimization problems have been well addressed during past decades, the continual lifelong learning scheme poses a great challenge, where the training samples are with a non-stationary distribution, corresponding to its task objective. 
	
	\begin{figure}[t!]
		\centering
		\includegraphics[width=0.9\columnwidth]{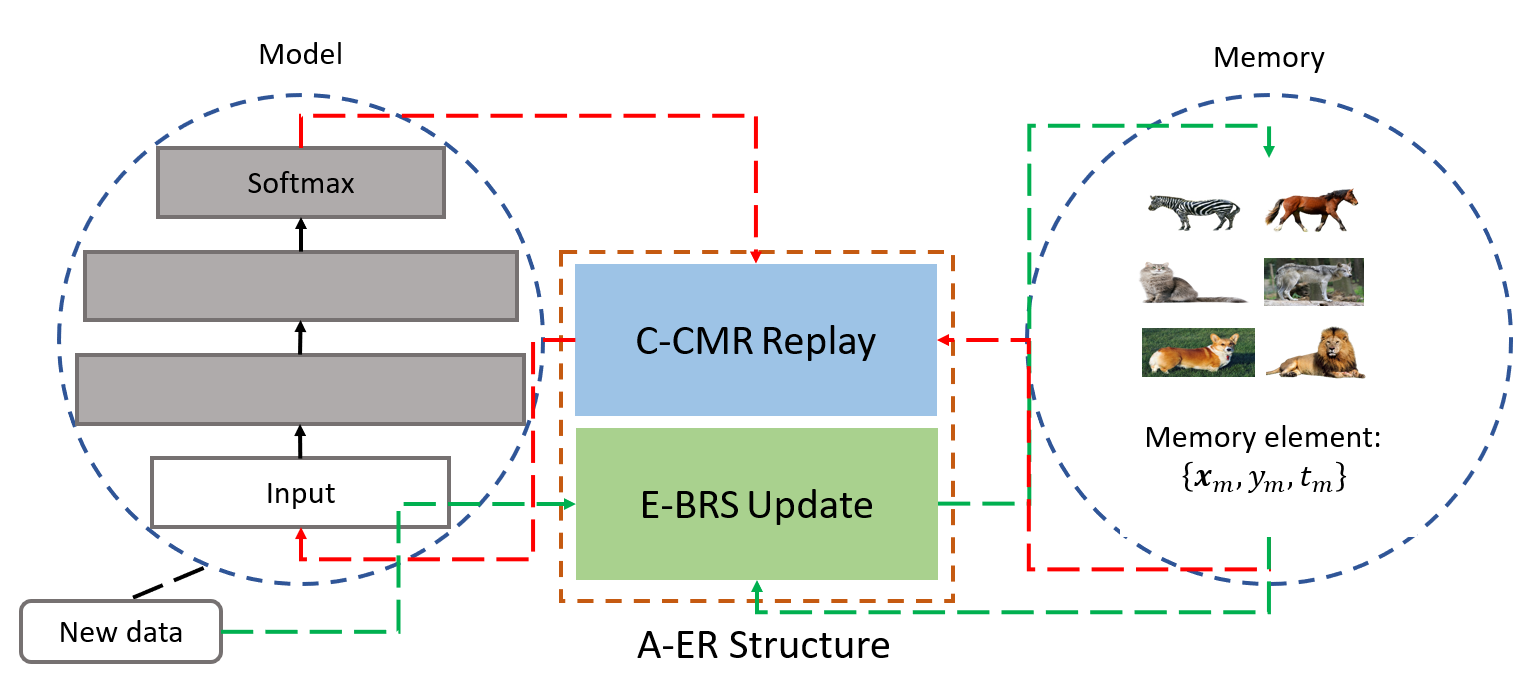}
		\caption{System diagram of the proposed AdaER: an adaptive experience replay algorithm with the developed contextually-cued memory recall (C-CMR) method for the replay stage and the entropy-balanced reservoir sampling (E-BRS) strategy for the update stage.}
		\label{fig:overview}
	\end{figure}
	In the continual lifelong learning setting, we have $\mathcal{X}$ divided into $T$ tasks, which is indexed as $\mathcal{X}_t \sim \mathcal{D}_t, t \in [1, \dots, T]$. Particularly, during the learning process of the $t$-th task, only the training examples from $\mathcal{X}_t$ can be accessible to the classifier $f$. Hence, the objective in Eq.~\eqref{Eq:stationary} cannot be learned directly. For better presentation on the learning of task $t$, we denote the distribution of previously seen training examples in $\mathcal{X}_s = \{\mathcal{X}_1, \dots, \mathcal{X}_{t-1}\}$ as $(\mathbf{x}_s, y_s) \sim \mathcal{D}_s$, where the objective of a ML learner to achieve Eq.~\eqref{Eq:stationary} can be described as:
	\begin{equation}\label{Eq:continual}
		\theta^{*}  = \argmin_{\theta} (\mathcal{L}_{t} + \mathcal{L}_{s}),
	\end{equation}
	where the first part $\mathcal{L}_{t}=  \mathbb{E}_{(\mathbf{x}_t, y_t)\sim \mathcal{D}_t} [l(f_\theta(\mathbf{x}_t), y_t)]$ requires the learner to rapidly learn the current task, and the second part $\mathcal{L}_{s}=  \mathbb{E}_{(\mathbf{x}_s, y_s)\sim \mathcal{D}_s} [l(f_\theta(\mathbf{x}_s), y_s)]$ denotes the requirement of not forgetting the previously knowledge. 
	Note that, during the learning of the current task, all training data are accessible with batches denoted by $\mathcal{B}_t \in \mathcal{X}_t$, which could be still viewed as identically drawn from $\mathcal{D}_t$. However, all training data of previously learned tasks are not available and it is difficult to obtain the $\mathcal{L}_{s})$ in Eq.~\eqref{Eq:continual}. Without the integration of $\mathcal{L}_{s}$, a typical ML learner will suffer from the catastrophic forgetting issue \cite{goodfellow2013empirical}, e.g., due to the lack of stability in neural networks \cite{kirkpatrick2017overcoming}. The requirement of minimizing both $\mathcal{L}_{t}$ and $\mathcal{L}_{s}$ is also known as the stability-plasticity dilemma in existing studies\cite{lopez2017gradient}. 
	
	\begin{figure*}[t!]
		\centering
		\includegraphics[width=0.8 \textwidth]{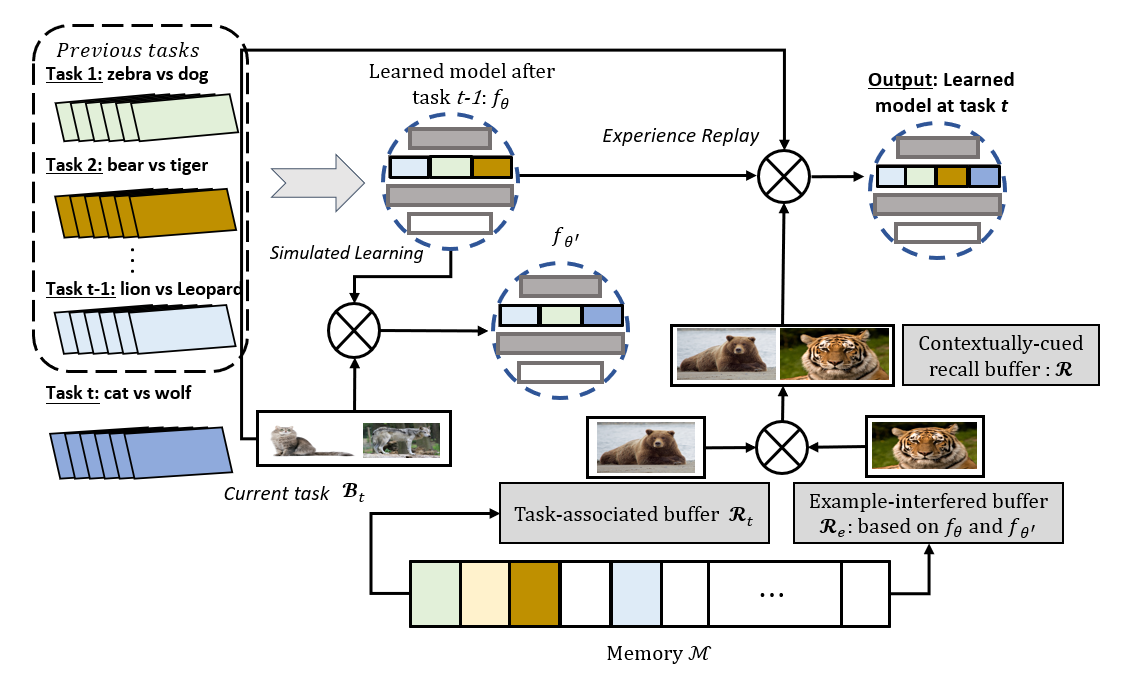}
		\caption{Illustration of the developed C-CMR method: the most contextually-cued memories $\mathcal{R}$ are replayed to mitigate the forgetting with the combination of example-interfered buffer $\mathcal{R}_e$ and task-associated buffer $\mathcal{R}_t$. }
\label{fig:structure}
\end{figure*}

To address this challenge, there have been several existing approaches in recent years, among which one popular method is known as experience replay (ER) \cite{rolnick2018experience}. The central feature of ER is to leverage a memory storage $\mathcal{M} \sim \mathcal{D}_m$ for previously seen training samples, where $ \mathcal{M} = \{(\mathbf{x}_1, y_1), \dots, (\mathbf{x}_m, y_m)\}$ and $|\mathcal{M}| = M$. Obviously, it is not realistic to either store every seen training samples in $\mathcal{M}$ or replay each example in $\mathcal{M}$ at every current learning step. Typically, given a limited size of memory, the mechanism of ER could be summarized into the following two steps: (i). \textit{Memory update:} $\mathcal{M}$ is updated when it learns more tasks, followed by a reservoir sampling method \cite{vitter1985random}. (ii). \textit{Memory replay:} during the training of the current task $t$, a batch of examples $\mathcal{B}_m = \{\mathbf{x}_m, y_m\}$ are randomly sampled from $\mathcal{M}$ that is interleaved with the current batch $\mathcal{B}_t$ to improve the stability of the learner. Specifically, during the learning of task $t$, ER approach addresses the objective in Eq.~\eqref{Eq:continual} as $\theta = \argmin_{\theta} (\mathcal{L}_{t} + \mathcal{L}_{m})$, where $\mathcal{L}_{m} =  \mathbb{E}_{(\mathbf{x}_m, y_m) \sim \mathcal{D}_m} [l(f_\theta(\mathbf{x}_m), y_m)]$. 

However, the performance of ER is affected by the distribution discrepancy between the replayed batch $\mathcal{B}_m$ and the previously seen data $\mathcal{D}_s$, which becomes more pronounced as memory resources become limited. Additionally, random reservoir sampling used for memory update can lead to imbalanced datasets, exacerbating the issue of catastrophic forgetting, especially for classes with fewer training examples.

To address these limitations, our proposed algorithm focuses on improving both memory replay and update. We aim to mitigate the distribution discrepancy between $\mathcal{B}_m$ and $\mathcal{D}_s$ and alleviate the imbalance issue in real-world scenarios. By addressing these challenges, we aim to enhance the performance of ER-based continual lifelong learning approaches.


\section{Methods}\label{Sec:method}
In this section, we present our novel continual lifelong learning algorithm, called Adaptive Experience Replay (AdaER). The structure of AdaER is depicted in Fig.~\ref{fig:overview}, and it addresses the limitations of existing experience replay (ER) methods by enhancing both the replay and update stages. In AdaER, the memory tuples in $\mathcal{M}$ store not only the previously seen training data, but also the corresponding task IDs. Each memory content in $\mathcal{M}$ is denoted as ${\mathbf{x}_m, y_m, t_m}$, where $\mathbf{x}_m$ represents the input sample, $y_m$ denotes the corresponding label, and $t_m$ indicates the task ID.

To improve the replay stage, AdaER introduces a Contextually-Cued Memory Recall (C-CMR) method (Sec.~\ref{Subsec:C_CMR}). C-CMR determines which memories should be replayed based on contextual cues, considering the interference caused by the data-conflicting and task-conflicting examples in the memory buffer. The goal is to select the most relevant memories for effective knowledge consolidation.

For the update stage, AdaER enhances the memory updating strategy by maximizing the corresponding information entropy. This strategy is known as Entropy-Balanced Reservoir Sampling (E-BRS) (Sec.~\ref{Subsec:E_BBRS}). By maximizing the information entropy, AdaER mitigates the imbalanced distribution issue that can arise in real-world scenarios, leading to improved performance and reduced forgetting.

The detailed design and implementation of AdaER are provided in the subsequent sections, which include the C-CMR method and the E-BRS strategy. These advancements aim to enhance the efficiency and effectiveness of ER-based continual lifelong learning methods, addressing the challenges associated with memory replay and update.



\subsection{Contextually-Cued Memory Recall}\label{Subsec:C_CMR}
In the C-CMR method, we illustrate its functionality using a continuous animal classification scenario depicted in Fig.~\ref{fig:structure}. The goal of C-CMR is to select appropriate memory examples for the seen animal classes. To achieve this, C-CMR employs two buffers: the example-interfered buffer $\mathcal{R}_e$ and the task-associated buffer $\mathcal{R}_t$. These buffers help determine the relevant memories from both the data-conflicting and task-conflicting perspectives. The selected memories are then stored in the contextually-cued buffer $\mathcal{R}$.

First, the C-CMR identifies the most interfered samples based on data conflicts and stores them in $\mathcal{R}_e$. These data-conflicting samples provide valuable information for knowledge consolidation. Simultaneously, C-CMR investigates the forgetting of associated tasks stored in $\mathcal{R}_e$ and identifies the task-related samples. These task-conflicting samples are then stored in $\mathcal{R}_t$. By considering both data conflicts and task conflicts, C-CMR leverages the information from both buffers, $\mathcal{R}_e$ and $\mathcal{R}_t$, to create the contextually-cued buffer $\mathcal{R}$.

The contextually-cued buffer $\mathcal{R}$ contains selected memory examples that are relevant for effective knowledge consolidation. By combining the information from the example-interfered buffer and the task-associated buffer, C-CMR ensures that the replayed memories are contextually appropriate and contribute to mitigating catastrophic forgetting in continual lifelong learning scenarios.

\subsubsection{Example-Interfered Buffer}
To identify memory examples that conflict with the current learning task, C-CMR introduces a virtual classifier $f_\theta'$, which is trained on the current batch $\mathcal{B}_t$ without any memory replay. This approach is inspired by previous work \cite{aljundi2019online} and addresses the stability-plasticity dilemma in lifelong learning. The learning of $f_\theta'$ at the $t$-th task is formulated as a one-step stochastic gradient descent (SGD) optimization problem, where the model parameters $\theta'$ are updated using the gradients computed on $\mathcal{B}_t$ with a learning rate $\alpha$. The update is performed according to the following equation:
\begin{equation}\label{Eq:virtual_theta}
\theta' = \theta - \alpha \nabla_\theta l(f_\theta; \mathcal{B}_t),
\end{equation}
where $\nabla_\theta l(\cdot)$ represents the gradient of the loss function $l(\cdot)$ with respect to the model parameters $\theta$. By updating $\theta$ using the gradients calculated on the current batch, the virtual classifier $f_\theta'$ is obtained.

The motivation behind using $f_\theta'$ is to assess the forgetting degree of each memory example with respect to the current learning task. Different memory examples exhibit varying degrees of forgetting, with some being transferable to the new task and others interfering or being forgotten. By comparing the performance of $f_\theta'$ and the original model $f_\theta$ on the memory examples, C-CMR quantifies the forgetting degree of each example, enabling the selection of conflicting memory examples for further analysis and handling. To quantize the forgetting degree of each memory example, we introduce a score vector $\mathbf{s}$ with a criterion, calculating $s(m)$ for the $m$-th memory sample as:
\begin{equation}\label{Eq:criterion}
s(m) = l(f_{\theta'} (\mathbf{x}_m), y_m ) - l (f_{\theta} (\mathbf{x}_m), y_m ),
\end{equation}
where $\mathbf{s} \in \mathbb{R}^{M \times 1}$. Note that when the value of $s(m)$ is higher, the $m$-th memory example is considered with a higher degree of forgetting against the learning of $\mathcal{B}_t$. As a result, C-CMR develops the example-interfered buffer $\mathcal{R}_e$ by choosing the top-$p$ interfered memories instead of randomly sampling from $\mathcal{M}$, where $p = |\mathcal{R}_e|$. 

\begin{figure}[t!]
\centering
\includegraphics[width=0.9 \columnwidth]{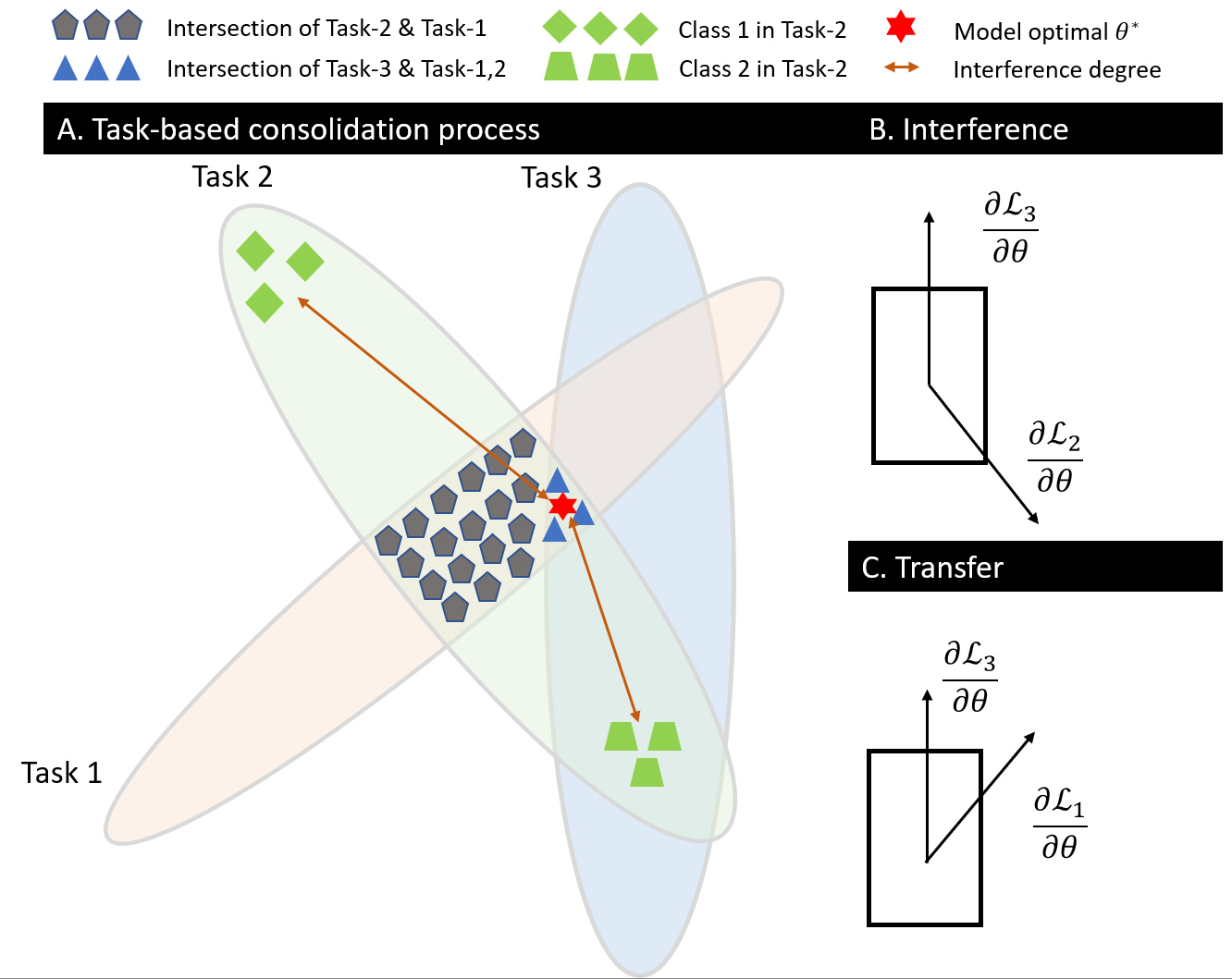}
\caption{Task-associated interference-transfer relationship with a three-task continual learning scenario.}
\label{fig:task}
\end{figure}
\subsubsection{Task-Associated Buffer}
Although the example-interfered buffer $\mathcal{R}_e$ captures the most representative forgotten examples from a data-conflicting perspective, it does not consider the relationship between transfer and interference among the learned tasks. This relationship is illustrated in a three-task continual learning example shown in Fig.~\ref{fig:task}. The objective in this scenario is to find the optimal model $\theta^\star$ after training on task 3, which lies in the overlapping region of all tasks. However, during the learning of task 3, there is a boundary situation where the memory of task 1 is transferred, while task 2 is interfered with. In this case, examples from class 1 in task 2 are more likely to be chosen for $\mathcal{R}_e$ as they have a larger distance to $\theta^\star$, resulting in a higher value of $s(m)$.

The presence of $\mathcal{R}_e$ can significantly reduce the learning performance of class 1, but it can also generate an offset to the optimal model $\theta^\star$. This offset may cause interference with class 2, which was learned together with class 1 in task 1. To address this issue, C-CMR introduces the task-associated buffer $\mathcal{R}_t$, which focuses on the forgetting of learned tasks from a task-conflicting perspective.

To distinguish between interfered and transferred tasks, the obtained memory examples in $\mathcal{R}_e$ are further analyzed. Let the task IDs in $\mathcal{R}_e$ be denoted as $j = [1, \dots, t_e]$, where $1 \leq t_e < p$. Considering the number of $j$-th task samples captured in $\mathcal{R}_e$ as $p_j$, the number of training memories from $\mathcal{M} - \mathcal{R}_e$ captured in $\mathcal{R}_t$ is defined as follows:

\begin{equation}\label{Eq:task_weight}
q_j = q \times \frac{p_j}{p}.
\end{equation}
Note that the obtained  $q_j$ can be viewed as a weighted factor of the $j$-th task in $\mathcal{R}_t$: when task $j$ is more interfered with during the continual learning process, the chances of its training samples being selected into $\mathcal{R}_e$ is higher, whose data distribution needs to pay more attention for offset correction.

\subsubsection{Contextually-Cued Recall}

\begin{algorithm}[t!]
\caption{C-CMR: A Contextually-Cued Memory Recall Approach for Continual Lifelong Learning}
\label{algorithm: alpha}
\begin{algorithmic}[1]
\STATE {\bfseries Input:} 
Initialized $f_{\theta}$, Memory $\mathcal{M}$, learning rate $\alpha$. 
\FOR {$t = 1 : T$}
\STATE Batch $\mathcal{B}_t$ for task $t$ is drawn from  $\mathcal{D}_t$.
\STATE $\theta^{'} = \theta -   \alpha \nabla_{\theta} l(f_{\theta} ;\mathcal{B}_t) $ as introduced  from Eq.~\eqref{Eq:virtual_theta}.
\IF {replay is true}
\STATE 	Develop $\mathbf{s}$ from $\theta^{'} $ and $\theta $ as in Eq.~\eqref{Eq:criterion}.
\STATE Search the most interfered examples $(\mathbf{x}_m, y_m)$ with highest values of $s(m)$ into $\mathcal{R}_e$.
\STATE Develop $\mathcal{R}_t$, where for the $j$-th task, the number of examples $q_j$ follows Eq.~\eqref{Eq:task_weight}.
\STATE $\mathcal{R} = \mathcal{R}_e + \mathcal{R}_t$.
\STATE$\theta = \theta - \alpha \nabla_{\theta} l(f_{\theta}; (\mathcal{R}+ \mathcal{B}_t)).$ \quad $\Rightarrow$ Update continual learner $\theta$ with the training of both current task batch and replay batch.  
\ENDIF			
\ENDFOR
\STATE 	Memory update $\mathcal{M}$ as introduced in Algorithm~\ref{algorithm: 1}.
\end{algorithmic}
\end{algorithm}

To address the challenge of transfer and interference in memory replay, our proposed C-CMR method combines both the example-interfered buffer $\mathcal{R}_e$ and the task-associated buffer $\mathcal{R}_t$ for replay. The final replay buffer obtained in C-CMR is denoted as $\mathcal{R} = \mathcal{R}_e + \mathcal{R}_t$. The memory replay process of C-CMR is summarized in Algorithm~\ref{algorithm: alpha}.

In order to ensure that the continual learner can learn the current task while replaying memories, we set the size of $\mathcal{R}$ and $\mathcal{B}_t$ to be the same. To further investigate the contributions of $\mathcal{R}_e$ and $\mathcal{R}_t$ in C-CMR, we introduce a new hyperparameter $\tau = p / (p+q)$, and its impact is studied in detail in Section~\ref{Sec:analysis}.


\subsection{Entropy-Balanced Reservoir Sampling}\label{Subsec:E_BBRS}
Meanwhile, the AdaER algorithm improves the memory update stage of ER-based methods by introducing the entropy-balanced reservoir sampling (E-BRS) method. This method aims to increase the diversity of training samples within the memory buffer $\mathcal{M}$ by maximizing its information entropy, which also addresses the issue of imbalanced data distribution.

In existing ER-based methods, random reservoir sampling is commonly used to select which training examples are stored in the fixed-size memory buffer $\mathcal{M}$. The probability of each training exemplar being represented in reservoir sampling is $M/N$, where $N$ is the number of seen samples. However, when the continual learning data is imbalanced, with some classes having fewer samples stored in $\mathcal{M}$, the risk of decreasing information entropy arises.

To address this, the E-BRS method encourages a balanced number of training samples for each class. This is achieved by approximating the information entropy through balancing the number of samples per class in $\mathcal{M}$, as shown in Algorithm~\ref{algorithm: 1}. This approximation reduces the computational overhead compared to estimating the information entropy using kernel functions. Additionally, E-BRS takes the criterion score $s(m)$ obtained from C-CMR into consideration. By removing the least important memory examples in Line~8, E-BRS prevents the most forgotten memory example from being replaced. This ensures that important information is retained in the memory buffer during the update stage.

\begin{algorithm}[t!]
\caption{Development of $M$ in AdaER: E-BRS}
\label{algorithm: 1}
\begin{algorithmic}[1]
\STATE {\bfseries Input:} 
	 Data pair $(\mathbf{x}, y)$ from $\mathcal{B}_t$, memory buffer $\mathcal{M}$, 
	seen examples $N$.
\IF{$M > N$ }
\STATE $\mathcal{M}[N] \leftarrow (\mathbf{x}, y)$.		
\ELSE
\STATE valid = RandInt$([0, N])$.
\IF {valid $\leq M $}
\STATE 	$\tilde{y} = \arg\max_{y} $ Count $(y \in \mathcal{M})$.
\STATE $m$ = $ \arg\min_{m} ( s(m) | y_m = \tilde{y} )$.
\STATE $\mathcal{M}[m] =  (\mathbf{x}, y)$
\ENDIF
\ENDIF
\STATE Updated memory buffer $\mathcal{M}$.
\end{algorithmic}
\end{algorithm}

\subsection{Discussions}


In this paper, we propose the AdaER algorithm to improve the efficiency of replay-based continual learning baselines. AdaER consists of two stages: C-CMR and E-BRS. The C-CMR method provides guidance for the replay strategy by considering both data-conflicting and task-conflicting examples. It combines the example-interfered buffer $\mathcal{R}_e$ and the task-associated buffer $\mathcal{R}_t$ to determine which memories to replay. The goal is to address catastrophic forgetting by replaying the most interfered memories. We compare our C-CMR method with the existing maximally interfered retrieval (MIR) approach and show that MIR fails in certain boundary scenarios where $\tau = 1$. We provide a detailed performance analysis of AdaER and MIR in Section~\ref{Sec:analysis}.

The E-BRS method improves the random reservoir sampling strategy used in the memory buffer updating process. Instead of estimating information entropy using kernel functions, we balance the number of samples per class in $\mathcal{M}$. This simplification avoids computational complexity concerns. 

Overall, AdaER combines the C-CMR and E-BRS methods to enhance the replay-based continual learning process. We show that AdaER outperforms existing baselines, including MIR, and provide detailed performance analysis in Section~\ref{Sec:analysis}.

\section{Experiments}\label{Sec:analysis}

\begin{table}[t!]
	\begin{adjustbox}{width=0.4\columnwidth, center}
		\begin{tabular}{*{4}{c}}
			\toprule
			Dataset  & $N_{train}$& $N_{task}$ & $N_{c}$  \\  
			\midrule
			split-MNIST \cite{lecun1998gradient} & $1,000$ & $5$ & $ 2$ \\
			split-FMNIST \cite{xiao2017fashion, farquhar2018towards} & $1,200$ & $5$  &$ 2$  \\
			split-CIFAR10 \cite{krizhevsky2009learning}& $10,000$ & $5$  & $ 2$ \\
			split-CIFAR100 \cite{krizhevsky2009learning} & $1,000$ & $50$  & $ 2$ \\
			\bottomrule
		\end{tabular}
	\end{adjustbox}
	\caption{Numerical Details of introduced benchmarks in this paper. For each benchmark, $N_{task}$ is the number of tasks, $N_{c}$ denotes the number of classes per task, and $N_{train}$ represents the training data size per each task respectively.}
	\label{Table:data}
\end{table}
\subsection{Experimental Setup}
To evaluate our proposed AdaER continual learning algorithm, we compared it with recently proposed baselines under the class-IL continual learning settings over several supervised benchmarks in \cite{van2019three, aljundi2019online}.

\subsubsection{Benchmarks}
In this work, we introduce the following benchmarks in the continual lifelong learning field, whose in-depth descriptions are summarized in  Table~\ref{Table:data}. Note that we rely on the {Split-MNIST}, {Split-FMNIST} and {Split-CIFAR10} benchmarks as the most representative continual learning tasks for performance analysis. And the {Split-CIFAR100} benchmark is used to test how the proposed AdaER algorithm performs on tasks with long sequences. 

\begin{table*}[t!]

\begin{adjustbox}{width=0.9\textwidth, center}
\begin{tabular}{*{13}{c}}
\toprule
\multicolumn{1}{c}{}&  \multicolumn{4}{c}{split-MNIST}  & \multicolumn{4}{c}{split-FMNIST}  & \multicolumn{4}{c}{split-CIFAR10}  \\
Method & Acc & Forget & Bwt & Fwt &Acc & Forget & Bwt & Fwt & Acc& Forget & Bwt & Fwt  \\ 
\midrule
Online & $18.4  $ &$ 98.4$ & $-78.4$ & N/A &$20.0 $ & $98.0$ & $-78.6$ &N/A& $13.0  $ &$ 83.6$ & $-66.8$ & N/A \\
Joint & $94.0  $ & N/A & N/A &  N/A & $83.2 $ & N/A &  N/A &  N/A & $63.6  $ & N/A& N/A &  N/A \\
\midrule
oEWC & $19.8$ & $98.9 $& $-98.7$ & $-14.4$ & $20.0 $ & $98.6$ & $-98.7$ & $-14.5$ & $17.7 $ & $78.4$ & $-58.3$ & $-12.9$  \\
SI & $19.4 $ & $ 99.2$  & $-98.8$ & $-13.4$ & $ 19.9 $ & $ 98.7$ & $98.8$ & $-13.3$ & $15.2 $ & $83.8$ & $-72.8$ & $-12.7$  \\
\midrule
\ ER &  $86.4  $ & $11.6 $ & $-10.7$ & $-7.07$ &  $69.6 $ & $23.6$ & $-18.5$ & $-6.89$  &  $34.0 $ & $39.0$  & $-19.9$ & $-12.5$   \\ 
GEM &  $76.7  $ & $22.9 $ & $-13.3$ & $-22.9$ &  $66.7 $ & $ 30.2$ & $-21.8$ & $-17.2$ &  $24.3 $ & $63.5 $ & $-17.3$ & $-18.9$ \\ 
A-GEM &  $38.7 $ & $67.1$ & $-57.0$ & $-14.7$ &  $32.8 $ & $65.8$ & $-70.5$ & $-12.8$ &  $19.3  $ & $74.5$ & $-43.7$ & $-12.6$  \\ 
iCaRL &  $70.3  $ & $14.7 $ & $-14.2$ & N/A  &  $65.0 $ & $29.5 $ & $-18.9$ & N/A  &  $33.1 $ & $50.0$ & $-24.4$ & N/A  \\ 
HAL & $84.2 $ & $ 19.4$  & $-12.4$ & $-11.6$ & $ 68.7 $ & $ 19.2$ & $-19.2$ & $-19.3$ & $31.8 $ & $43.8$ & $-33.7$ & $-12.9$  \\
GSS & $85.6 $ & $ 14.3$  & $-8.9$ & $-10.7$ & $ 69.2 $ & $ 22.7$ & $-16.5$ & $-22.3$ & $42.3 $ & $29.6$ & $-21.7$ & $-13.3$  \\
MIR &  $88.0  $ & $9.0$  & $-8.9$ & $-7.07$ &  $71.4 $ & $22.4 $ & $-16.9$ & $-6.89$ &  $44.6 $ & $25.0 $& $-0.8$ & $-12.5$  \\ 
C-CMR &  $88.4  $ & $7.2$ & $-6.6$ & $-7.01$ &  $73.0$ & $18.5$ & $-14.6$ & $-6.87$ &  $45.4 $ & $22.6$ & $3.5$ & $-12.5$  \\
E-BRS &  $88.6  $ & $7.0$ & $-6.0$ & $-7.01$ &  $72.9$ & $21.2$ & $-14.8$ & $-6.88$ &  $45.2 $ & $24.2$ & $2.6$ & $-12.5$  \\
\textbf{AdaER} &  \bm{$89.6$} & \bm{$6.6$}  & \bm{$-5.4$} & \bm{$-6.90$} &  \bm{$74.0$} & \bm{$18.0$} & \bm{$-13.2$} & \bm{$-6.78$} &  \bm{$46.2$} & \bm{$18.0$} & \bm{$4.4$} & \bm{$-12.4$} \\ 
\bottomrule
\end{tabular}
\end{adjustbox}
\caption{Continual learning results for Split-MNIST, Split-FMNIST, and Split-CIFAR10. We report the mentioned four metrics: Acc (higher is better), Forget (lower is better), Bwt (higher is better), and Fwt (higher is better). Note that the results are spitted into different categories via the horizontal lines: the auxiliary joint and online baseline, the regularization-based methods, and the memory-based experience-replay baselines.}
\label{Table:analysis}
\end{table*}

\subsubsection{Baselines}
We compare the proposed AdaER algorithm with the following existing baselines in recent literature: \textbf{oEWC} \cite{schwarz2018progress}, \textbf{SI} \cite{zenke2017continual}, \textbf{GEM} \cite{lopez2017gradient}, \textbf{AGEM} \cite{chaudhry2019tiny}, \textbf{iCaRL} \cite{rebuffi2017icarl}, \textbf{ER} \cite{rolnick2018experience}, \textbf{MIR} \cite{aljundi2019online}, \textbf{GSS},. and \textbf{HAL} \cite{chaudhry2020using}. 
Moreover, we also evaluate the performance of the following two different settings. \textbf{Online:} the learner is trained under the continual learning setting by simply applying SGD optimizer. \textbf{Joint:} all tasks are trained jointly as one complete dataset instead of in a continual manner, which usually gives us an upper bound of the learning performance of all tasks.

\subsubsection{Training}
To provide a fair comparison with existing baselines, we train all the neural networks in this paper with the Stochastic Gradient Descent (SGD) optimizer. Additionally, all compared benchmarks in the paper are executed with the same computational resources. For MNIST and FMNIST benchmarks, we use a two-layer MLP with $400$ hidden nodes, which follows the settings in \cite{lopez2017gradient, riemer2018learning}. For CIFAR-10 and CIFAR-100, we use a standard Resnet-18 \cite{he2016deep} which is introduced in \cite{rebuffi2017icarl}. Note that for the replay-based methods, the batch size for $\mathcal{B}_t$ and $\mathcal{R}$ are both set to $20$, and the memory buffer is set to $100$ by default.

%
%

\subsubsection{Metrics}
In this paper, we measure the performance of continual learning algorithms with the following four metrics, which are defined in the literature \cite{chaudhry2019tiny, lopez2017gradient}. Note that for the $T$ tasks in a continual learning benchmark, we evaluate the test performance after learning each task. As such, we conduct a result matrix $R \in \mathbb{R}^{T \times T}$, where $R_{i,j}$ denotes the testing accuracy of the continual learner on task $t_j$ after learning task $t_i$. Let $\bar{b}_i$ be the testing accuracy for each task after the random initialization of the learning model and $F_i$ be the best testing accuracy for task $t_i$, the introduced four metrics are as follows. Note that for Acc, BWT, and FWT, the higher value indicates better performance, the lower the better for Forget. 
\begin{itemize}
\item $\textbf{Average Accuracy: } \text{Acc} = \frac{1}{T}\sum_{i=1}^{T}R_{T,i} $.
\item $\textbf{Average Forgetting: } \text{Forget} = \frac{1}{T-1}\sum_{i=1}^{T-1} R_{T,i} - F_i $.
\item $\textbf{Backward Transfer: } \text{Bwt} = \frac{1}{T-1}\sum_{i=1}^{T-1} R_{T,i} - R_{i,i} $.
\item $\textbf{Forward Transfer: } \text{Fwt} = \frac{1}{T-1}\sum_{i=1}^{T-1}R_{i,i} -\bar{b}_i $.
\end{itemize}

%
%
%

\subsection{Results}

\begin{figure*}[t!]
\centering
\begin{subfigure}{0.3\columnwidth}
\includegraphics[width = 1\columnwidth]{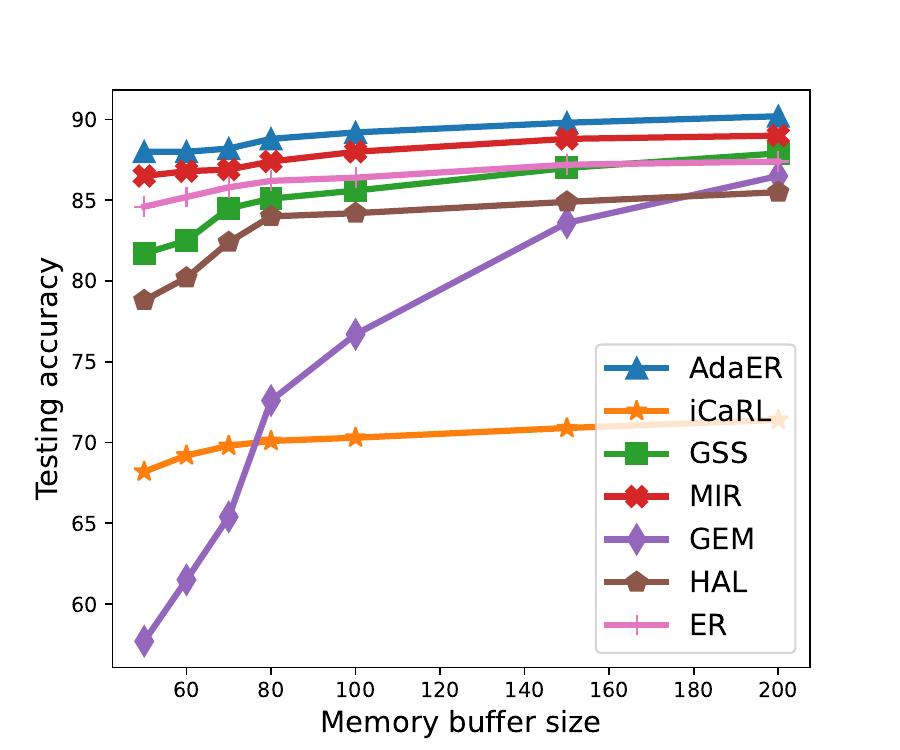}
\caption{split-MNIST}
\label{fig:mnist_mem_acc}
\end{subfigure}
\begin{subfigure}{0.3\columnwidth}
\includegraphics[width = 1\columnwidth]{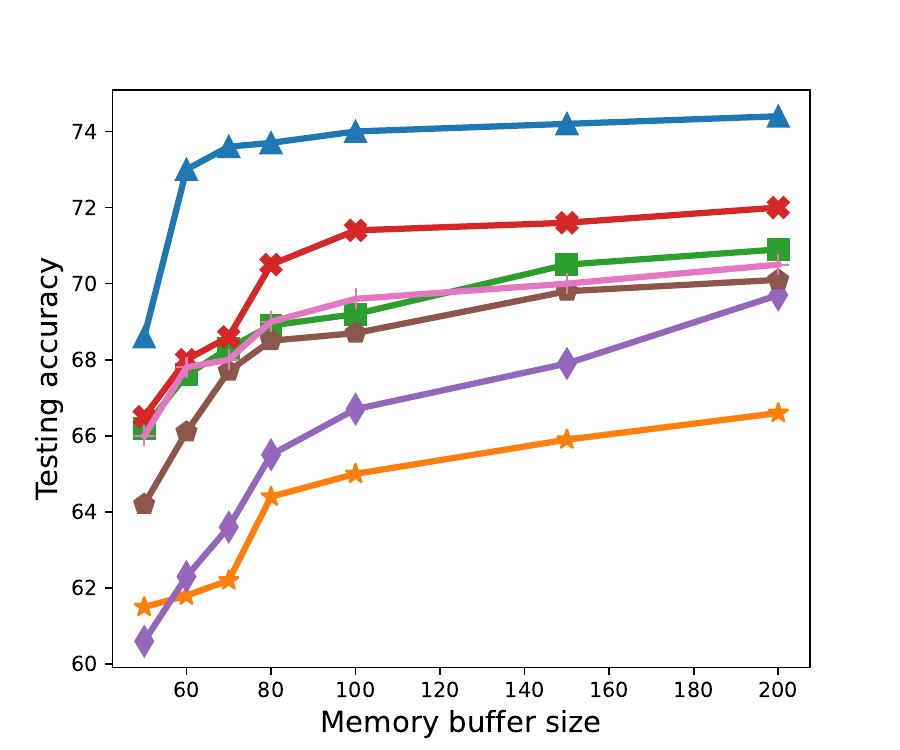}
\caption{split-FMNIST}
\label{fig:fmnist_mem_acc}
\end{subfigure}
\begin{subfigure}{0.3\columnwidth}
\includegraphics[width = 1\columnwidth]{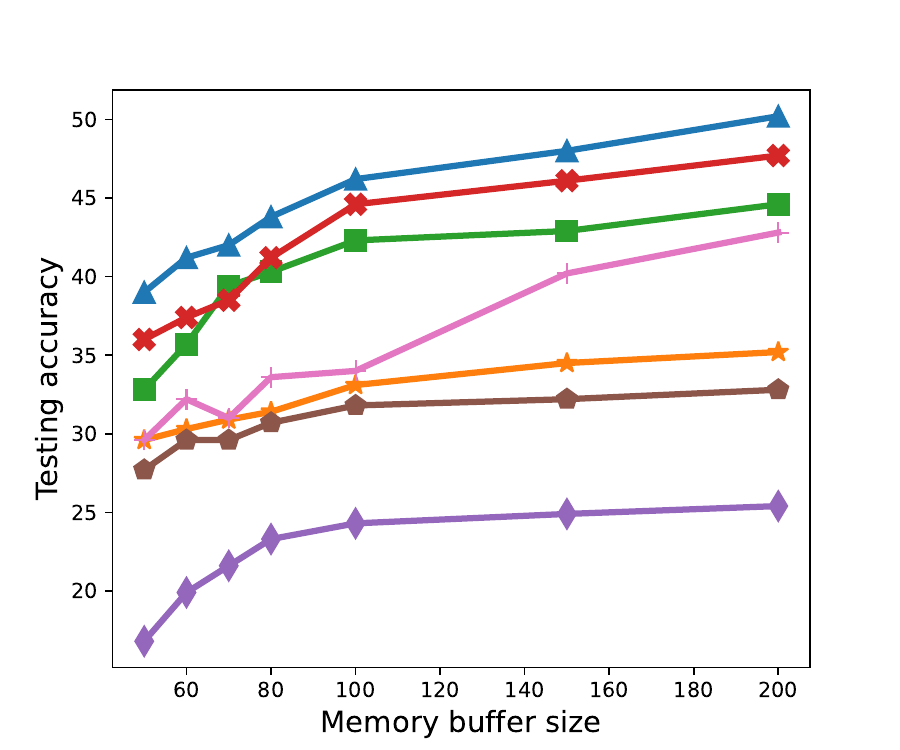}
\caption{split-CIFAR10}
\label{fig:cifar_mem_acc}
\end{subfigure}
\caption{The impact of different memory size over averaged testing accuracy.}
\label{fig:mem_acc}
\end{figure*}

\textbf{Performance analysis.} Table.~\ref{Table:analysis} shows the performance of compared continual learning baselines against three benchmarks, which are demonstrated with the four evaluation metrics. Note that for all memory-based approaches, the size of the memory buffer is $100$, and the continual learner trains each incoming training batch only once during the learning process of each benchmark. We can notice that the proposed AdaER algorithm achieves the best overall performance against every compared method at each introduced evaluation metric. 

Specifically, for split-MNIST and split-FMNIST, AdaER achieves $89.6\%$ and $74.0\%$ testing accuracy, which is $3.7\%$ and $6.3\%$ higher than the ER method respectively. For the split-CIFAR10 benchmark, AdaER obtains a positive backward transfer result as $4.4$, while the Bwt of ER is only $-19.9$. Additionally, compared to the MIR method, the forgetting of the proposed AdaER for split-CIFAR10 is only $18.0$, which is $28.0\%$ lower. And for the forward transfer, we can notice that compared to GSS, the proposed AdaER achieves $-6.78$ on the split-FMNIST benchmark which is $69.5\%$ higher. 

Furthermore, we also notice several interesting phenomenons: i) through GEM obtains feasible performance on split-MNIST and split-FMNIST, it performs poorly on split-CIFAR10, which may indicate GEM has limited ability against complex continual learning tasks; ii) though the averaged testing accuracy of GSS is very close to MIR method, it has worse performance on backward transfer, especially on split-FMNIST and split-CIFAR10 benchmarks. 
\begin{figure*}[t!]
\centering
\begin{subfigure}{0.3\columnwidth}
\includegraphics[width = 1\columnwidth]{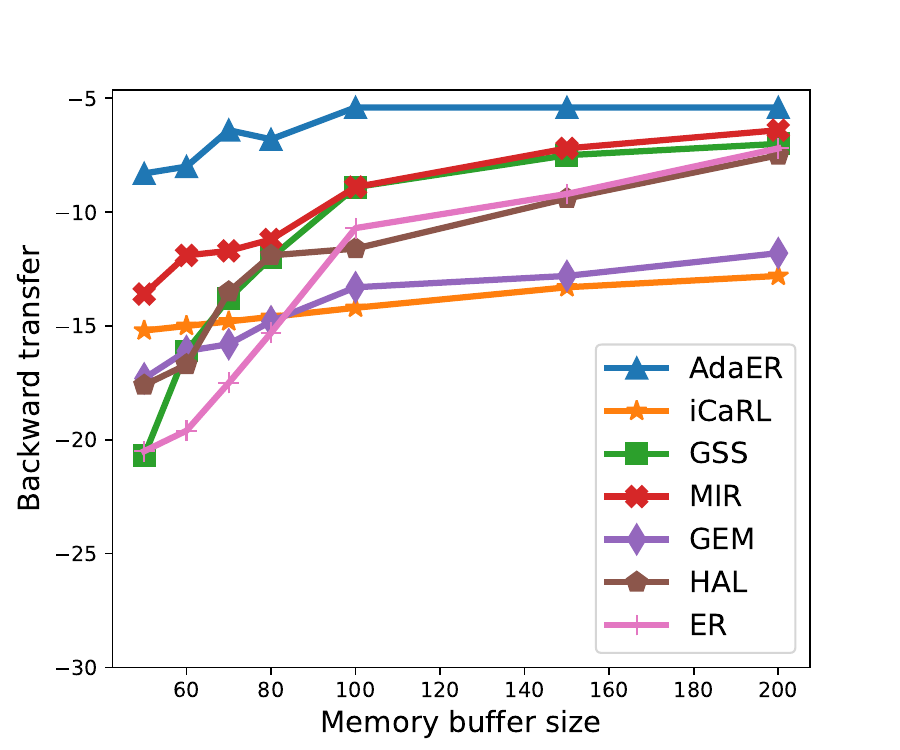}
\caption{split-MNIST}
\label{fig:mnist_mem_transfer}
\end{subfigure}
\begin{subfigure}{0.3\columnwidth}
\includegraphics[width = 1\columnwidth]{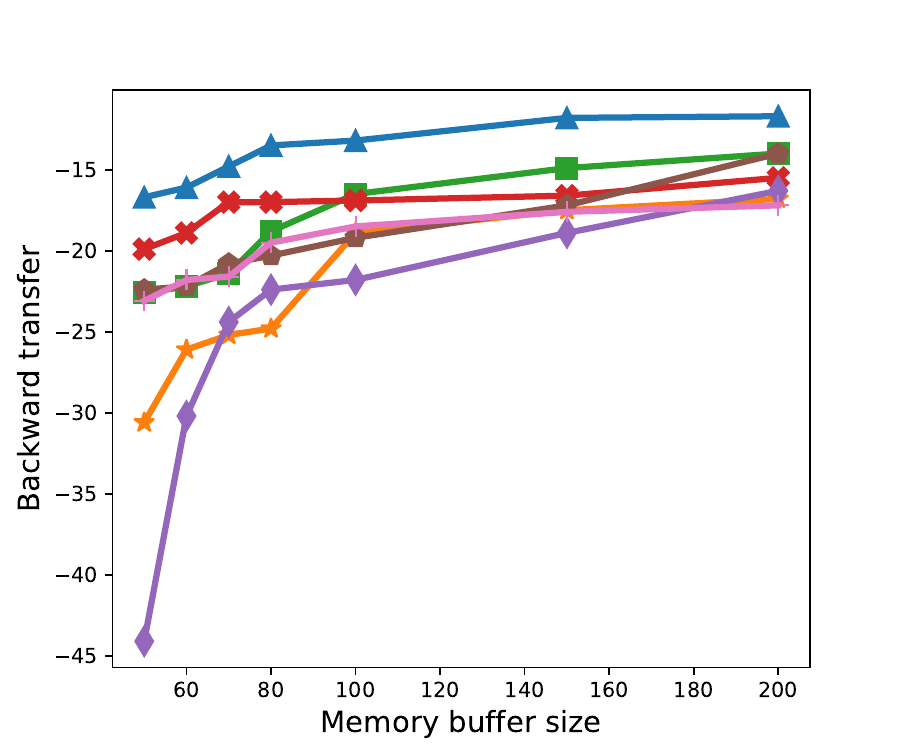}
\caption{split-FMNIST}
\label{fig:fmnist_mem_transfer}
\end{subfigure}
\begin{subfigure}{0.3\columnwidth}
\includegraphics[width = 1\columnwidth]{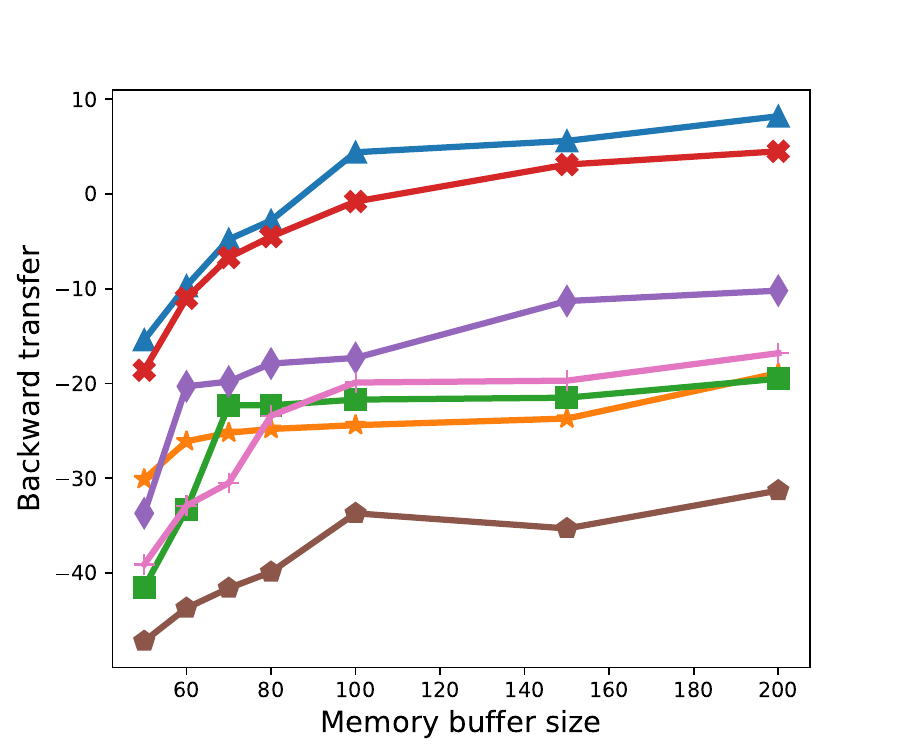}
\caption{split-CIFAR10}
\label{fig:cifar_mem_transfer}
\end{subfigure}
\caption{The impact of different memory size over backward transfer.}
\label{fig:mem_transfer}
\end{figure*}

\textbf{Impacts of memory capacity.} We then study the importance of the memory buffer size by evaluating the compared methods under different values of $M \in [50, 200]$. Firstly, we investigate the performance of compared baselines under different $M$ in terms of the testing accuracy, where the results are shown in Figure.~\ref{fig:mem_acc}. It can be noticed that as $M$ increases, the performance of compared continual learning methods becomes better, and the testing accuracy of the proposed AdaER algorithm outperforms other baselines on all benchmarks. Furthermore, in split-MNIST, the testing accuracy of GEM increases by $49.9\%$ when $M$ increases from $50$ to $200$, while only $2.5\%$ in AdaER. This indicates that the proposed AdaER algorithm is more robust against the memory capacity $M$, compared to other existing continual learning approaches.

\begin{figure*}[t!]
\centering
\begin{subfigure}{0.3\columnwidth}
\includegraphics[width = 1\columnwidth]{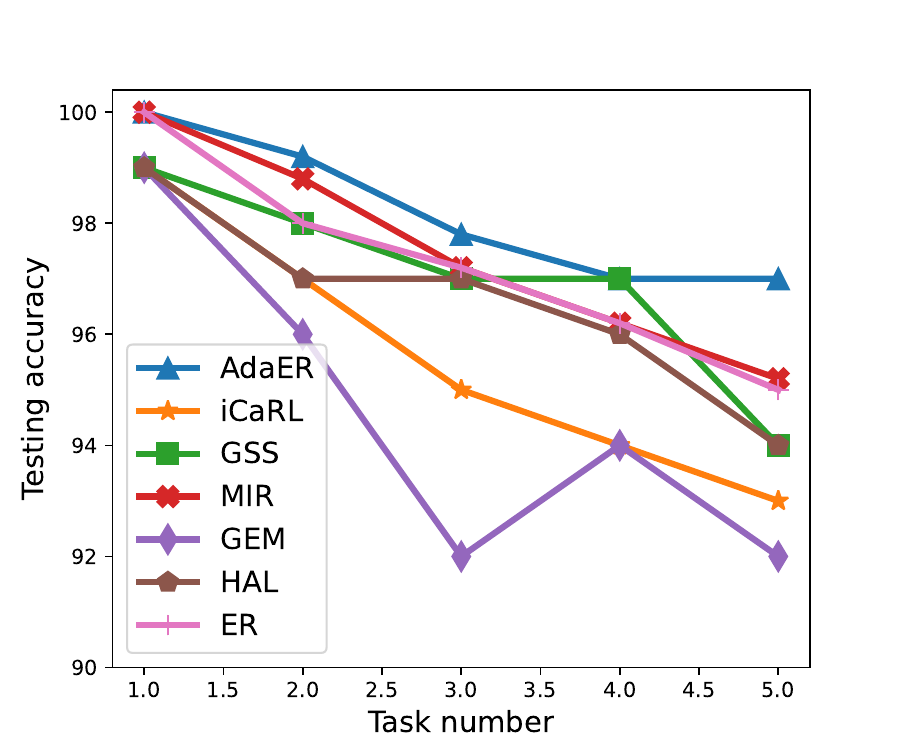}
\caption{split-MNIST}
\label{fig:mnist_mem_task}
\end{subfigure}
\begin{subfigure}{0.3\columnwidth}
\includegraphics[width = 1\columnwidth]{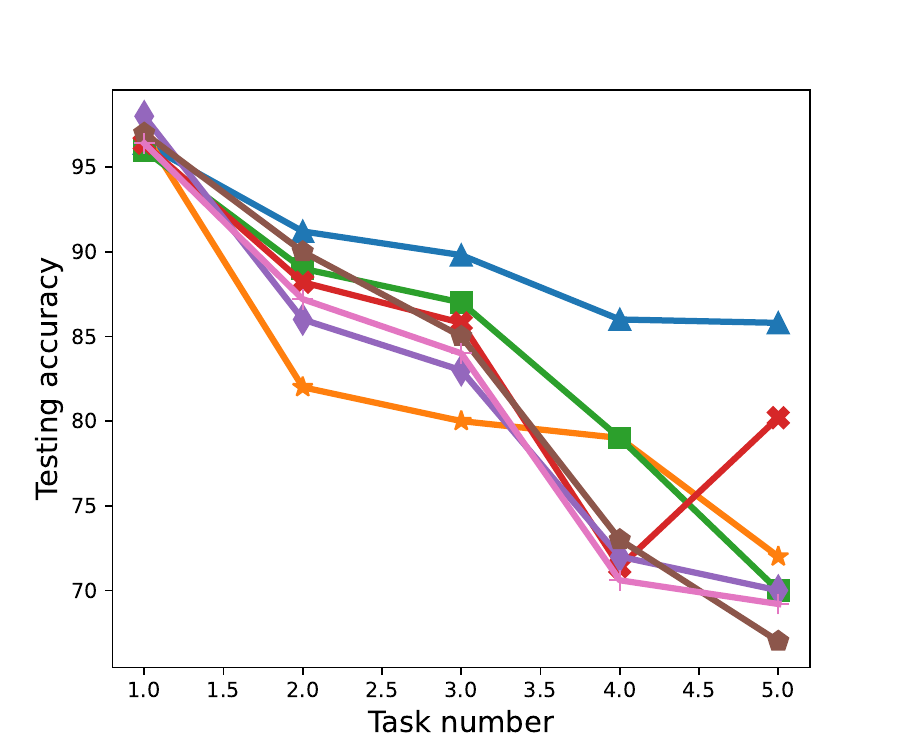}
\caption{split-FMNIST}
\label{fig:fmnist_mem_task}
\end{subfigure}
\begin{subfigure}{0.3\columnwidth}
\includegraphics[width = 1\columnwidth]{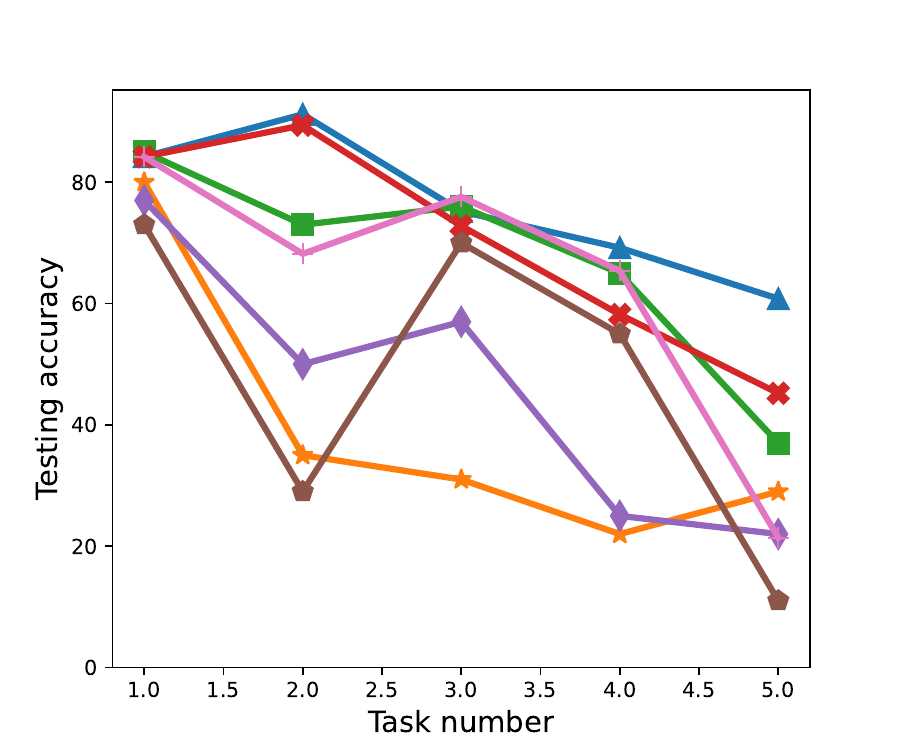}
\caption{split-CIFAR10}
\label{fig:cifar_mem_task}
\end{subfigure}
\caption{The testing accuracy of the first task as other tasks are learned.}
\label{fig:mem_task}
\end{figure*}

The results in  Figure.~\ref{fig:mem_transfer} demonstrate the changes in backward transfer performance with different memory buffer sizes. Similar to the results in Figure.~\ref{fig:mem_acc}, as the size of $\mathcal{M}$ increases, the proposed AdaER consistently outperforms all compared continual learning baselines in terms of both backward transfer and robustness. For example, in split-CIFAR10, the backward transfer value of the proposed AdaER increases from $-15.4$ to $8.2$, which outperforms other methods significantly.

\begin{figure*}[t!]
\centering
\begin{subfigure}{0.5\columnwidth}
\includegraphics[width = 1\columnwidth]{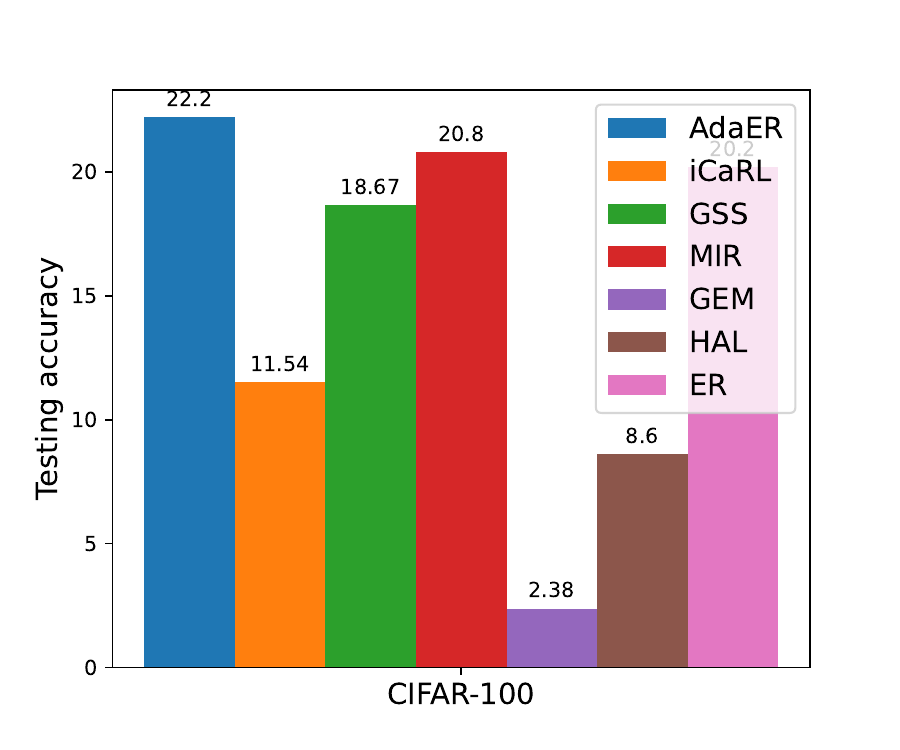}
\caption{Averaged testing accuracy}
\label{fig:cifar100_acc}
\end{subfigure}
\begin{subfigure}{1\columnwidth}
\includegraphics[width = 1\columnwidth]{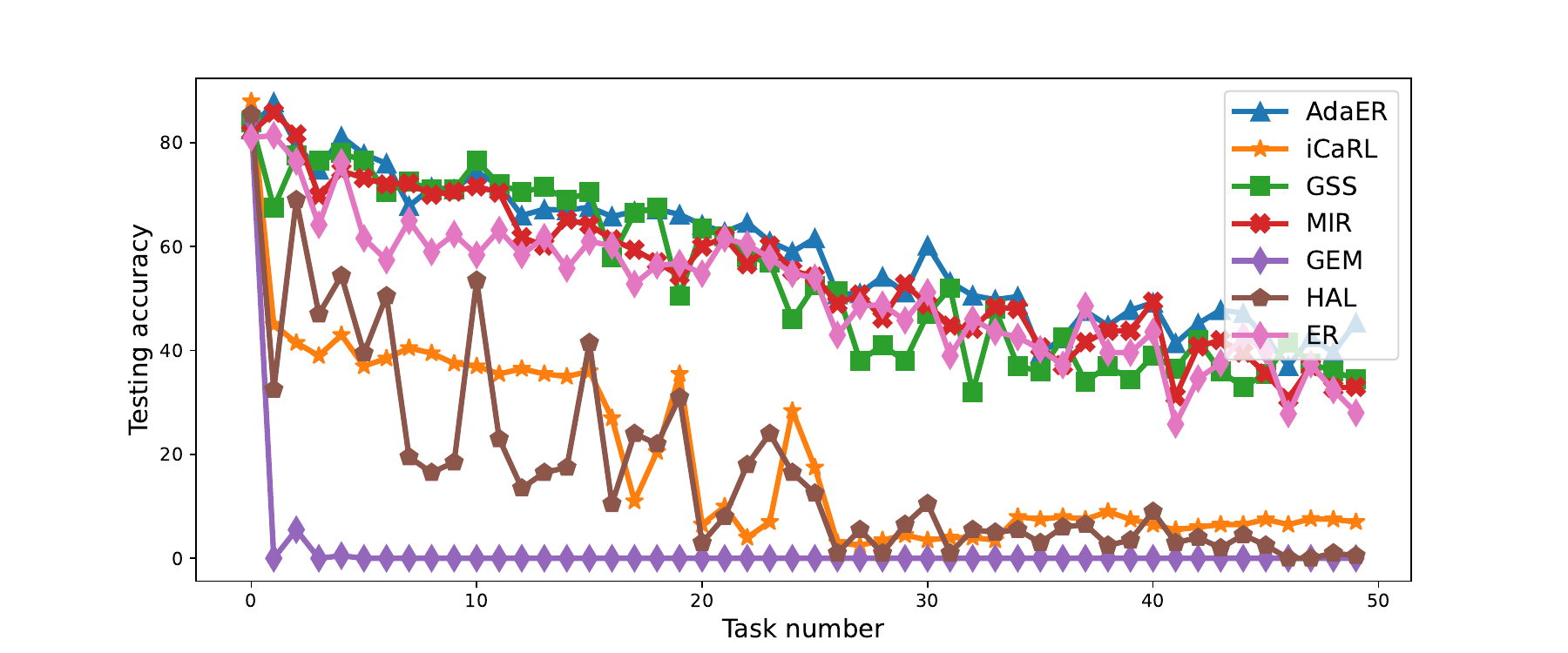}
\caption{Evolution of the first task}
\label{fig:cifar100_task}
\end{subfigure}
\caption{Results of the compared baselines against split-CIFAR100 benchmark.}
\label{fig:cifar_100}
\end{figure*}

\begin{figure*}[t!]
\centering
\begin{subfigure}{0.3\columnwidth}
\includegraphics[width = 1\columnwidth]{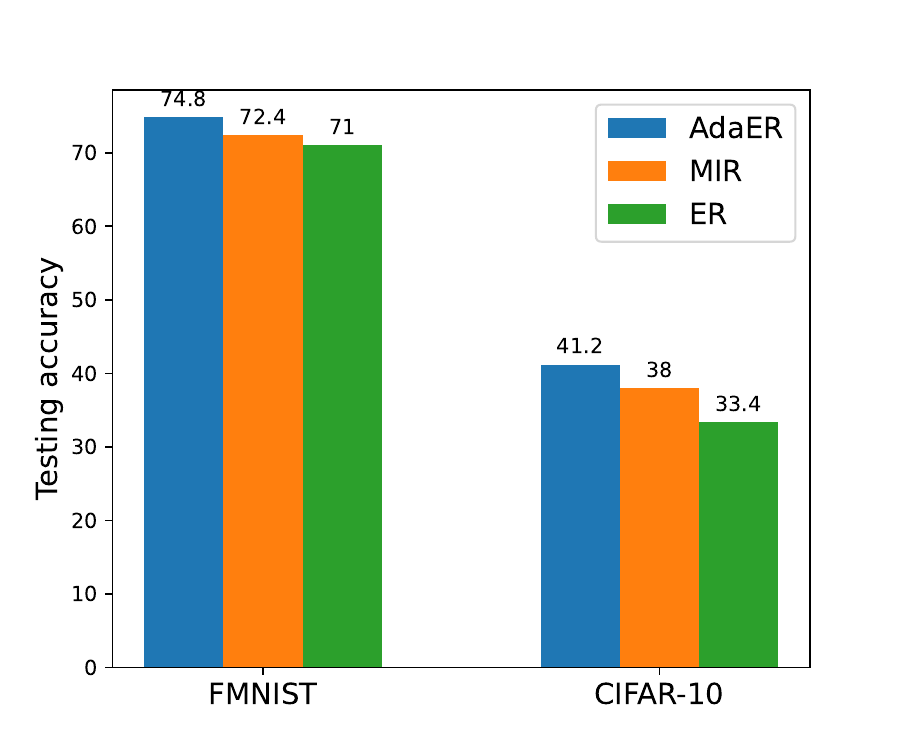}
\caption{Averaged testing accuracy}
\label{fig:imb_bar}
\end{subfigure}
\begin{subfigure}{0.3\columnwidth}
\includegraphics[width = 1\columnwidth]{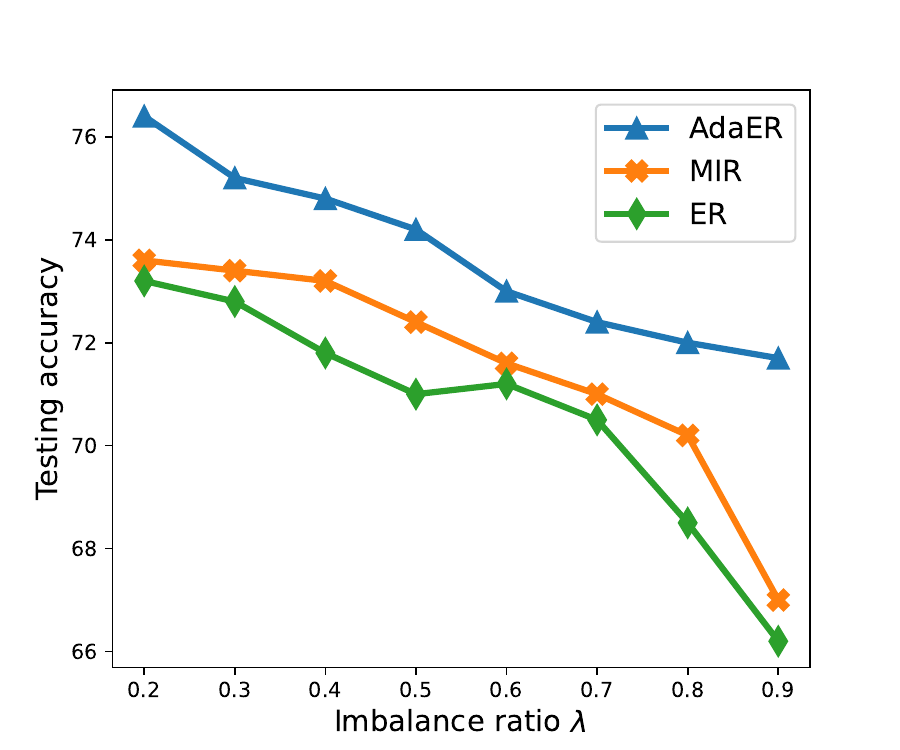}
\caption{split-FMNIST}
\label{fig:imb_fmnist_acc}
\end{subfigure}
\begin{subfigure}{0.3\columnwidth}
\includegraphics[width = 1\columnwidth]{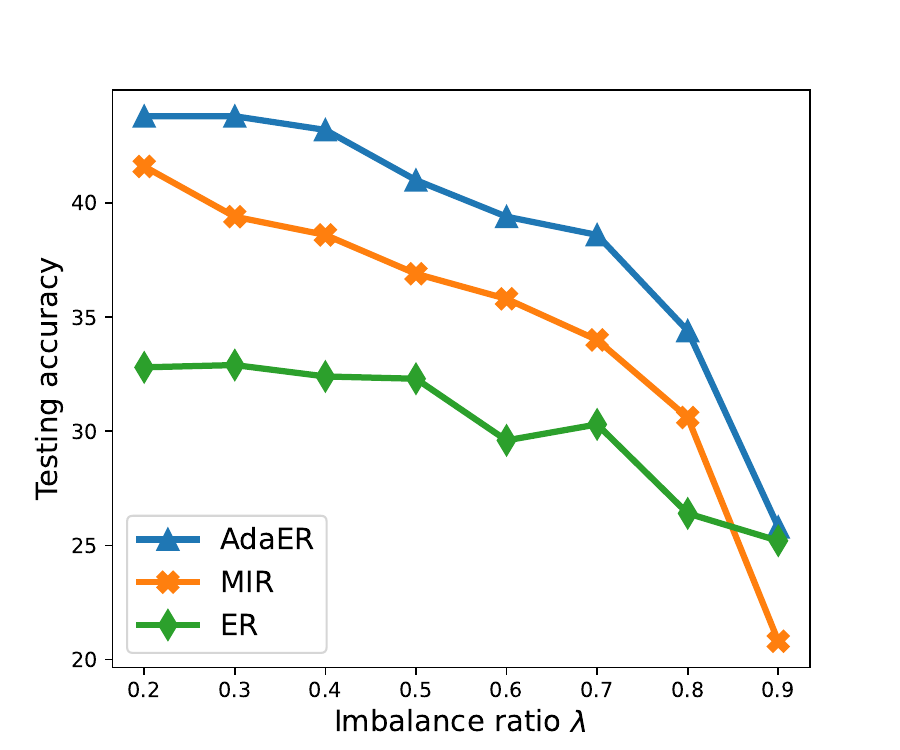}
\caption{split-CIFAR10}
\label{fig:imb_cifar_acc}
\end{subfigure}
\caption{Performance of the compared approaches under different settings of imbalanced training dataset partition: when $\lambda$ is larger, the imbalance degree is higher. }
\label{fig:imbalance_cifar}
\end{figure*}

\textbf{Investigation of the first task.} In Figure.~\ref{fig:mem_task}, we study the change of the testing accuracy of the first task in each benchmark throughout the complete continual learning process, i.e., how the forgetting evolves with more tasks being learned. The results show that compared to all baselines, the proposed AdaER algorithm achieves the overall minimal forgetting of the performance of the first task for each benchmark. For example, the first task accuracy of AdaER decreases $3\%$, $10.6 \%$ and $23.4\%$ corresponding to split-MNIST, split-FMNIST, and split-CIFAR10, which is $61.5\%$, $35.8\%$ and $40\%$ better than the results of MIR method in this work.

\begin{figure*}[t!]
\centering
\begin{subfigure}{0.3\columnwidth}
\includegraphics[width = 1\columnwidth]{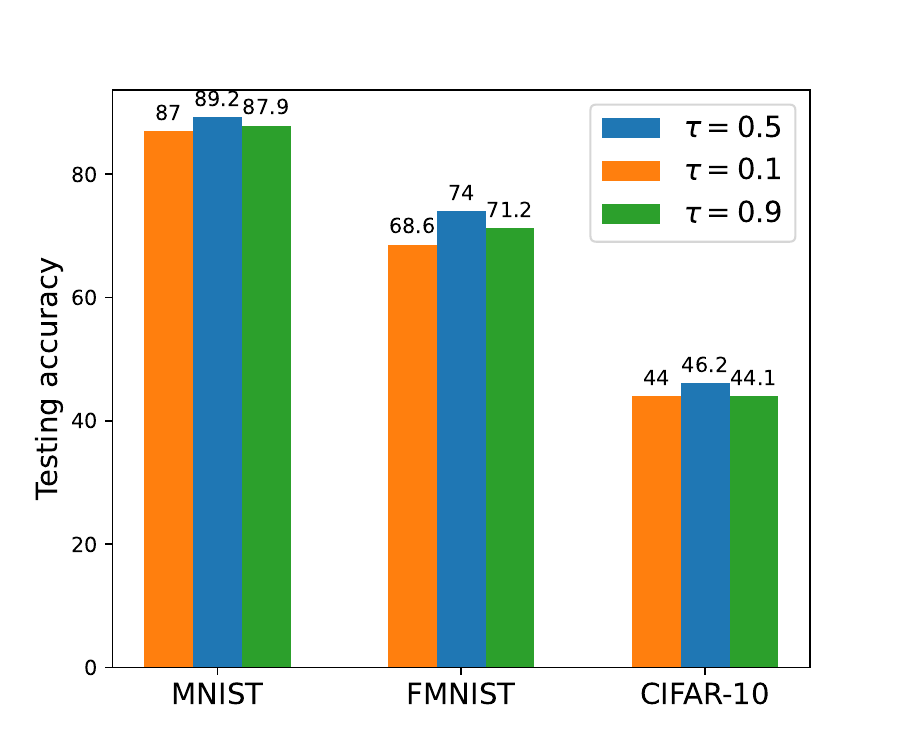}
\caption{Averaged testing accuracy}
\label{fig:trade_cifar_acc}
\end{subfigure}
\begin{subfigure}{0.3\columnwidth}
\includegraphics[width = 1\columnwidth]{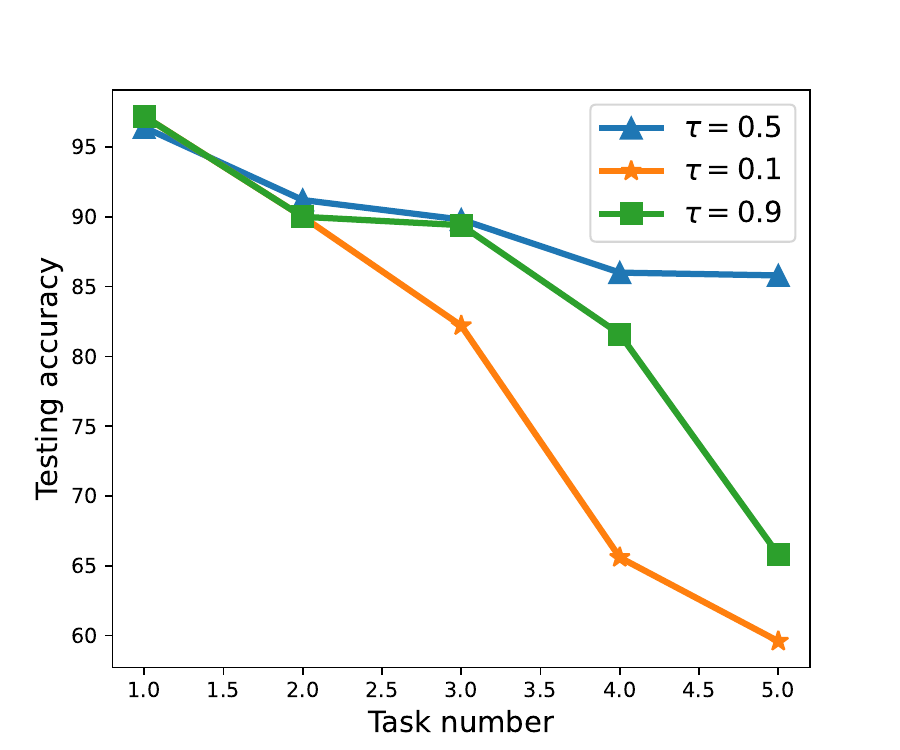}
\caption{split-FMNIST}
\label{fig:trade_cifar_mem}
\end{subfigure}
\begin{subfigure}{0.3\columnwidth}
\includegraphics[width = 1\columnwidth]{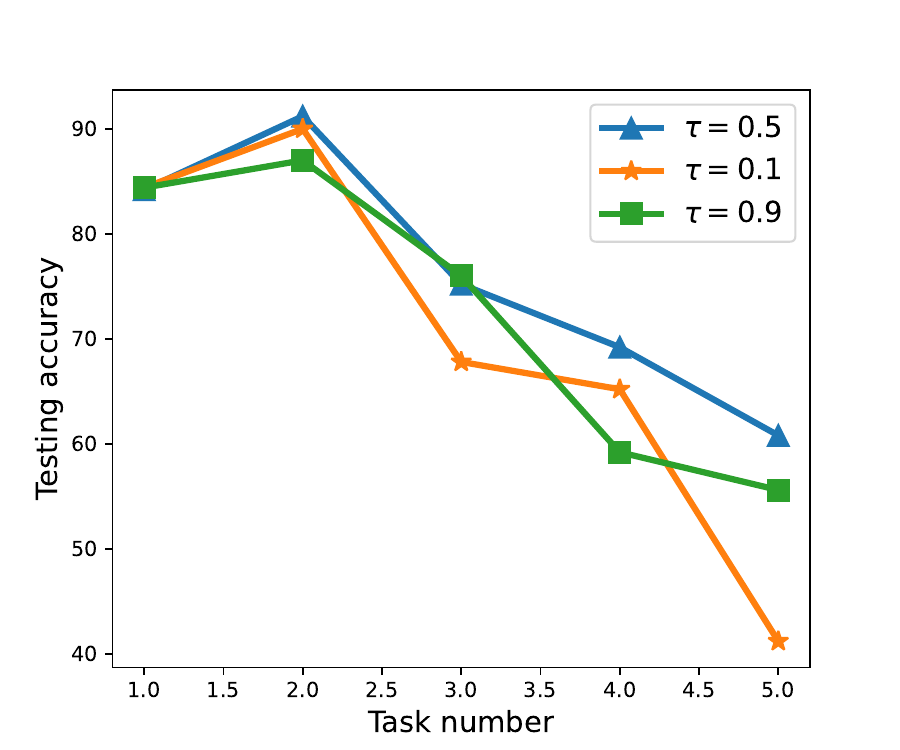}
\caption{split-CIFAR10}
\label{fig:trade_cifar_task}
\end{subfigure}
\caption{Performance of the proposed AdaER algorithm under different settings of $\tau$. (a): the averaged testing accuracy for each benchmark; (b): the evolution of first task testing accuracy on split-FMNIST; (c): the evolution of first task testing accuracy on split-CIFAR10. }
\label{fig:trade_cifar}
\end{figure*}

In addition, we also notice that the evolution of the first task is not monotonically decreasing. For example, the testing accuracy of AdaER after observing task $2$ is significantly higher than task $1$, which indicates that the proposed AdaER algorithm is with feasible transfer ability. 

\textbf{Study of long sequence task.} We then test the performance of the proposed AdaER algorithm on continual learning tasks with longer sequences. To achieve this, we conduct the split-CIFAR100 benchmark, where the training data of $100$ different labels are spitted into $50$ tasks, $2$ classes per each task. Note that this benchmark is extremely challenging in our experimental settings. As such, different from other benchmarks that each training batch is seen by the learner only once during the learning process, we add the iterations per each batch to $5$ for split-CIFAR100. The results in Figure.~\ref{fig:cifar_100} show the performance of compared approaches against this benchmark, where the left one is the averaged testing accuracy and the right one illustrates the evaluation of testing accuracy of the first task. 

It can be noticed from the results in Figure.~\ref{fig:cifar100_acc} that GEM, iCaRL, and HAL methods perform poorly and the proposed AdaER outperforms ER, GSS, and MIR baselines against split-CIFAR100. Compared to ER with $20.2\%$ testing accuracy, our AdaER algorithm achieves $22.2\%$ which is $9.9\%$ higher. Additionally, the first task evolution results in Figure.~\ref{fig:cifar100_task} also support our claims on GEM, iCaRL, and HAL methods, e.g., the testing accuracy of the first task in the GEM method drops rapidly to nearly zero after task $3$. Specifically,  while MIR and ER only have $33\%$ and $29\% $ final testing accuracy of the first task, the AdaER achieves $45.4\%$, which is $37.6\%$ and $56.6\%$ higher correspondingly. Besides, we can also notice that compared to ER, though the GSS method achieves a higher final first task testing accuracy, the averaged testing accuracy for every continual learning task in split-CIFAR100 is lower.

\begin{figure*}[t!]
\centering
\begin{subfigure}{0.3\columnwidth}
\includegraphics[width = 1\columnwidth]{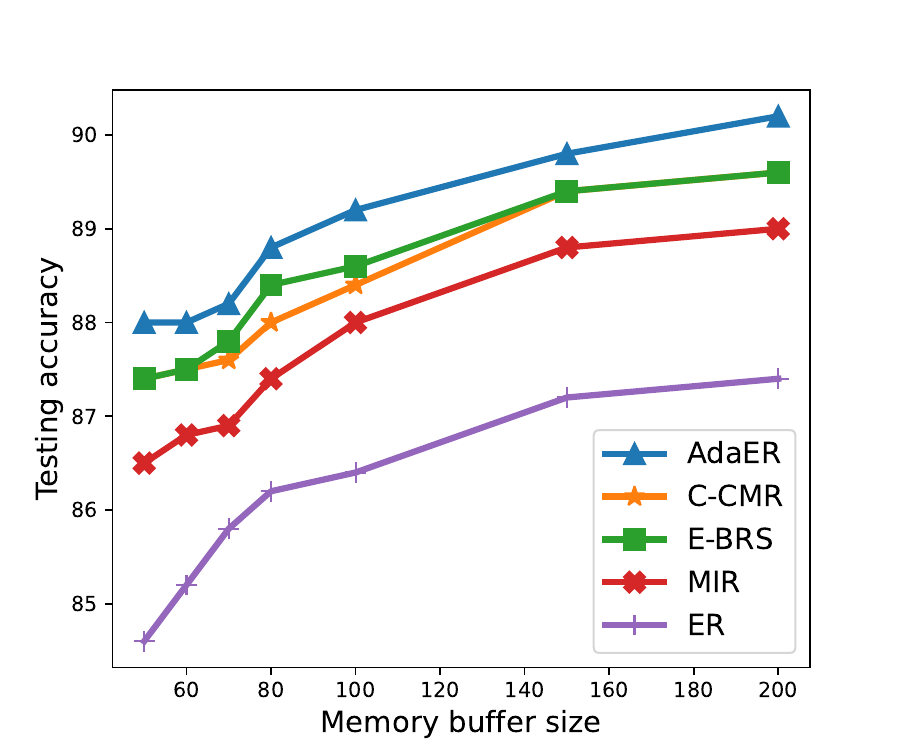}
\caption{split-MNIST}
\label{fig:ablation_mnist_mem_task}
\end{subfigure}
\begin{subfigure}{0.3\columnwidth}
\includegraphics[width = 1\columnwidth]{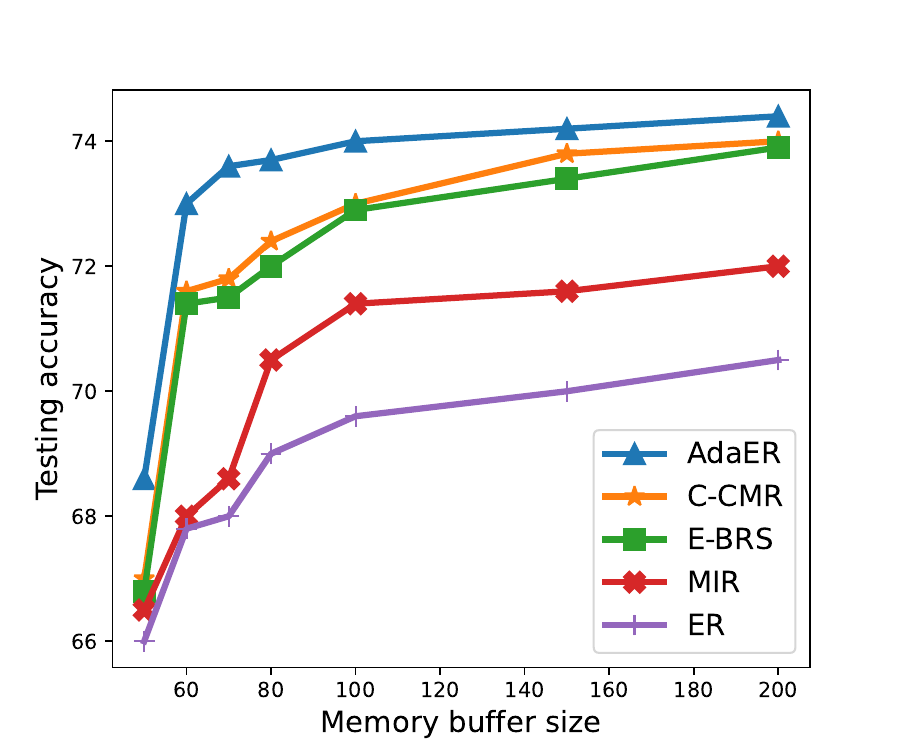}
\caption{split-FMNIST}
\label{fig:ablation_fmnist_mem_task}
\end{subfigure}
\begin{subfigure}{0.3\columnwidth}
\includegraphics[width = 1\columnwidth]{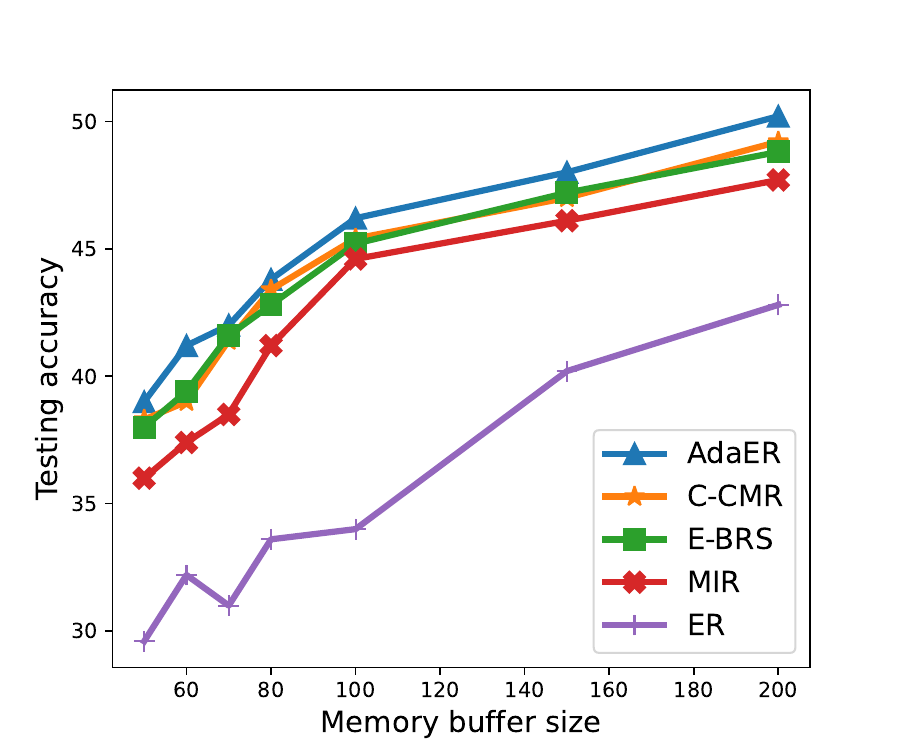}
\caption{split-CIFAR10}
\label{fig:ablation_cifar_mem_task}
\end{subfigure}
\caption{Ablation study of the proposed AdaER algorithm with the developed C-CMR and E-BRS methods: the performance of averaged testing accuracy as the increase of memory buffer size $M$.}
\label{fig:ablation_mem_task}
\end{figure*}

\begin{figure*}[t!]
\centering
\begin{subfigure}{0.5\columnwidth}
\includegraphics[width = 1\columnwidth]{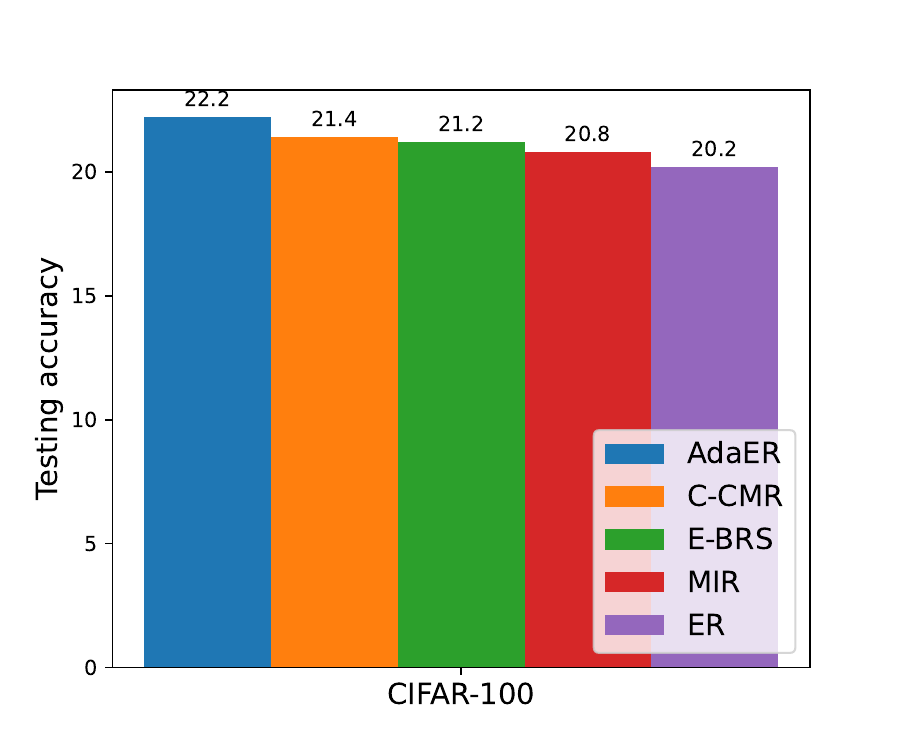}
\caption{Averaged testing accuracy}
\label{fig:ablation_cifar100_acc}
\end{subfigure}
\begin{subfigure}{1\columnwidth}
\includegraphics[width = 1\columnwidth]{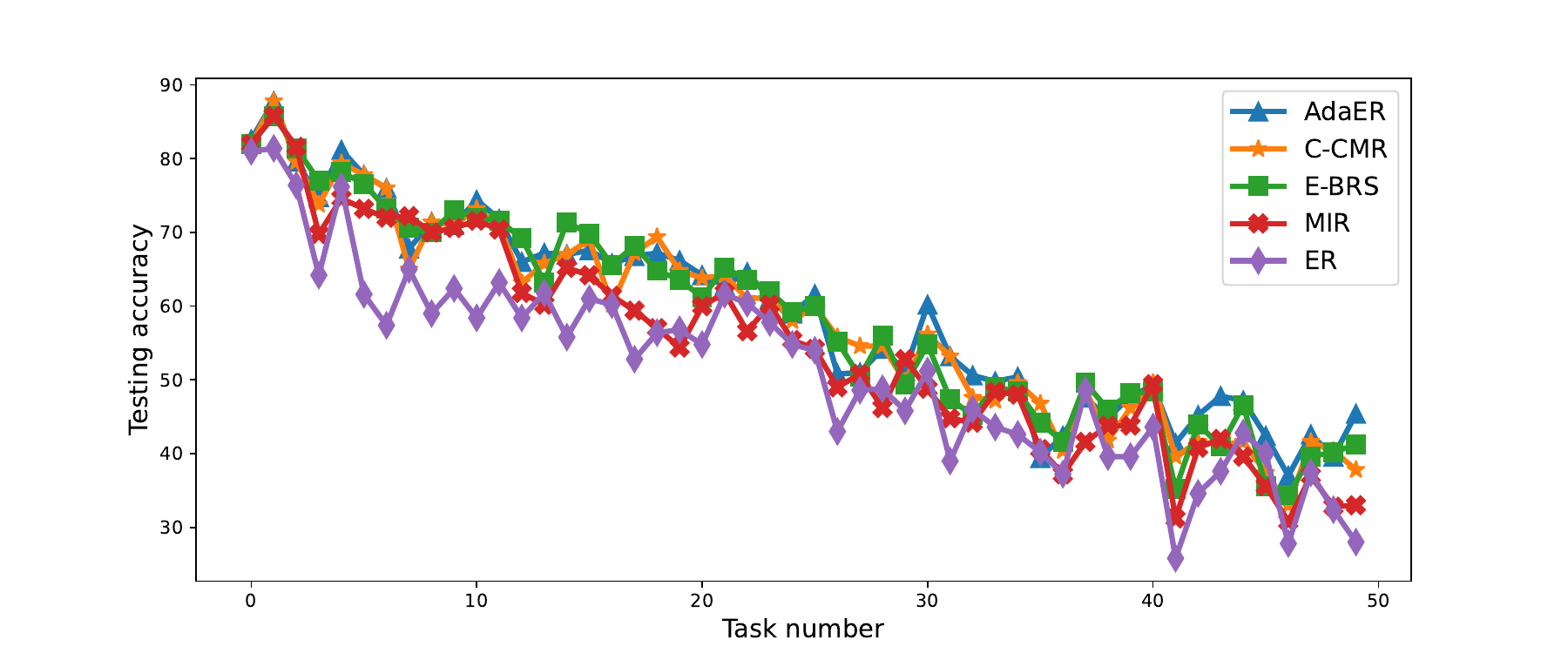}
\caption{Evolution of the first task}
\label{fig:ablation_cifar100_task}
\end{subfigure}
\caption{Ablation study of the proposed AdaER algorithm with the developed C-CMR and E-BRS methods over split-CIFAR100.}
\label{fig:ablation_cifar_100}
\end{figure*}

\textbf{Impacts of $\lambda$.} Then, we study the impact of imbalanced training data partition on the proposed AdaER algorithm. Note that for better evaluation, the performance of compared ER and MIR methods are also introduced for comparison. We setup the imbalanced data partition scenario on split-FMNIST and split-CIFAR10 benchmarks: for each task, the number of training samples in class $1$ to class $2$ is randomly set to $\lambda$. 

We first provide the averaged testing accuracy results of compared methods in Figure.~\ref{fig:imb_bar}, where $\lambda = 0.5$. From the results, we can notice that the proposed AdaER algorithm outperforms both MIR and ER under the imbalanced scenario. Compared to the results in Table.~\ref{Table:analysis}, all three introduced methods have a significant testing accuracy decrease against the split-CIFAR10 dataset. In this condition, we further evaluate this performance decrease and find that the proposed AdaER achieves minimal decline. The testing accuracy decline of AdaER is only $10.8\%$ while $14.8\%$ on MIR and $14.3\%$ on ER, which supports our claims in Section.~\ref{Subsec:E_BBRS}. 

Additionally, we investigate the performance changes of compared baselines with $\lambda \in [0.1, 0.9]$, of which the testing accuracy results are shown in Figure.~\ref{fig:imb_fmnist_acc} for split-FMNIST and \ref{fig:imb_cifar_acc} for split-CIFAR10. It can be noticed that as the value of $\lambda$ increases, the performance of each compared method decreases significantly. However, the proposed AdaER algorithm also shows its robustness to the imbalanced data partition. For example, in split-FMNIST, the testing accuracy decline from $\lambda = 0.9$ to $0.1$ is $6.2\%$, $9.0\%$ and $9.6\%$ for AdaER, MIR and ER respectively. And for split-CIFAR10, though the performance of AdaER drops faster than ER, it still outperforms ER over testing accuracy for each considered $\lambda$. 

\textbf{Impacts of $\tau$.} In this part, we evaluate the influence of the introduced hyper-parameter $\tau \in (0,1)$ on the performance of the proposed AdaER algorithm. As illustrated in Section.~\ref{Subsec:C_CMR}, $\tau$ denotes the weighted factor of task-associated buffer $\mathcal{R}_t$ and example-interfered buffer $\mathcal{R}_e$ that $\tau = p \slash |\mathcal{R}|$. Particularly, when $\tau = 0.5$, we consider $q$ is equal to $p$. We first provide the averaged testing accuracy with different values of $\tau$ in Figure.~\ref{fig:trade_cifar_acc}, where the results show that when $\tau = 0.5$, the proposed AdaER algorithm achieves the best among other settings. For example, in split-FMNIST, when $\tau = 0.5$, the averaged testing accuracy of AdaER is $74.0\%$, which is $7.9\%$ and $3.9\%$ higher than  $\tau = 0.1$ and $\tau = 0.9$ respectively. 

Additionally, we study the evolution of the first task testing accuracy with different values of $\tau$ on split-FMNIST and split-CIFAR10, whose corresponding results are shown in Figure.~\ref{fig:trade_cifar_mem} and \ref{fig:trade_cifar_task}. Firstly, the results also support that when $\tau=0.5$, the performance of AdaER is the best. Specifically, for example, in split-CIFAR10, the final testing accuracy of the first task when $\tau = 0.5$ is $60.8\%$, which is $47.6\%$ higher against $\tau=0.1$ and $9,4\%$ higher against $\tau=0.9$.

Then, we also notice that compared to the performance of $\tau=0.1$, the first task testing accuracy is more robust when $\tau =0.9$. We consider this phenomenon comes from the following two reasons. i) when $|\mathcal{R}|$ is fixed and $\tau$ is small, the size of the example-interfered buffer is too small, which may have limited ability to target the most forgotten tasks in the continual learning process; ii) when $\tau$ is close to $1$, the proposed AdaER shares the same feature with MIR, which aims to protect learning memory of the most forgot examples.

\subsection{Ablation Study}
In this part, we provide the ablation study of the proposed AdaER algorithm, where the two methods C-CMR and E-BRS are evaluated independently. Particularly, we evaluate C-CMR via the proposed contextually cued recall method and the memory buffer is developed with a random reservoir sampling strategy. And for E-BRS we develop the memory buffer with entropy-balanced reservoir sampling with the MIR replay strategy. Note that for better presentation, the experimental results of ER and MIR methods are also introduced for comparison. 

\textbf{Study of memory buffer size $M$.} We conduct the ablation study under the balanced training data partition setting as $\lambda = 0$. The results of averaged testing accuracy for the compared methods under different memory buffer sizes are shown in Figure.~\ref{fig:ablation_mem_task}. From the results, we can find that both C-CMR and E-BRS outperform MIR and ER at each benchmark. Interestingly, the performance of E-BRS is slightly better than C-CMR against split-MNIST while on the contrary against split-FMNIST. Particularly, as shown in Table.~\ref{Table:analysis}, when $M=100$, the averaged testing accuracy of C-CMR and E-BRS are $45.4\%$ and $45.2$, which are $33.5\%$ and $32.9\%$ higher than the ER method. 



\textbf{Study of long sequence task.} At last, we provide the ablation study of C-CMR and E-BRS methods against the most challenging split-CIFAR100 benchmark in Figure.~\ref{fig:ablation_cifar_100}. As shown in Figure.~\ref{fig:ablation_cifar100_acc}, compared to MIR and ER, both the proposed C-CMR and E-BRS achieve a higher averaged testing accuracy as $21.4\%$ and $21.2\%$, that are $5.9\%$ and $5.0\%$ higher against ER in detail. Additionally, we also provide the evolution of the first task testing accuracy in Figure.~\ref{fig:ablation_cifar100_task}. The results show that both C-CMR and E-BRS have a more robust first task testing accuracy against MIR and ER. Specifically, the final testing accuracy of E-BRS is $41.2\%$, which is $24.8\%$ higher than MIR and $42.1\%$ higher than ER.

\section{Conclusions}
The catastrophic forgetting problem is a long-standing challenge in the study of continual lifelong learning, especially in class-IL scenarios. While recently developed experience replay approaches have shown promising capability in mitigating this problem, their performance is still limited by its weakness of randomly sampling strategies on both the replay and update stages. As such, in this work, we propose the adaptive experience replay (AdaER) algorithm, which improves the two stages of existing ER via two methods. For the replay stage, AdaER provides a novel contextually-cued recall strategy, which considered both the interfered examples and the associated tasks during continual learning process that guides which of the memory examples should be replayed. For the update stage, we develop entropy-balanced reservoir sampling (E-BRS), which improves the original reservoir sampling strategy by maximizing the information entropy of the memory buffer. The experimental results show that the proposed AdaER algorithm outperforms existing approaches against class-IL continual learning.

The AdaER algorithm presented in this paper offers considerable advancements in the catastrophic forgetting problem in Experience Replay based class-IL lifelong learning scenario. We believe it can be incorporated into existing machine learning models to enhance their ability to retain and utilize knowledge from earlier learning stages, thereby improving their overall performance and adaptability. Additionally, the contextually-cued recall strategy and entropy-balanced reservoir sampling can offer significant improvements to the process of continual machine learning model training. 

There are several promising directions to explore in our future work. Firstly, further research could investigate how to optimize the parameters of the AdaER algorithm to achieve better performance by investigating implicit relationship between the example-interfered and the task-associated buffer. Additionally, while the current version of AdaER focuses on class-IL continual learning, it would be interesting to explore its applicability and performance in other continual learning scenarios, such as task-IL and domain-IL. Finally, the entropy-balanced reservoir sampling strategy proposed in this work could inspire other entropy-based strategies to enhance the efficiency of continual learning approaches.

\section{Acknowledgement}

This research was partially funded by US National Science Foundation (NSF), Award IIS 2047570.

\bibliographystyle{elsarticle-num}  
\bibliography{main_arxiv}

%

\noindent\textbf{Xingyu Li}
received the B.S. degree from the School of Electronic Science and Engineering (National Model Microelectronics College), Xiamen University, Xiamen, China, in 2015, and the M.S. degree from the Department of Electrical and Computer Engineering, Stevens Institute of Technology, Hoboken, NJ, USA, in 2017. He is currently pursuing the Ph.D. degree with the Department of Electrical and Computer Engineering, Mississippi State University, MS, MS, USA. His current research interests include federated learning and continual learning.\\
%
\noindent\textbf{Bo Tang} is an Assistant Professor in the Department of Electrical and Computer Engineering at Mississippi State University. He received the Ph.D. degree in electrical engineering from University of Rhode Island (Kingstown, RI) in 2016. From 2016 to 2017, he worked as an Assistant Professor in the Department of Computer Science at Hofstra University, Hempstead, NY. His research interests lie in the general areas of statistical machine learning and data mining, as well as their various applications in cyber-physical systems, including robotics, autonomous driving and remote sensing. \\
%
%
\noindent\textbf{Haifeng Li} received his Master degree in Transportation Engineering from South China University of Technology, Guangzhou, China, in 2005, and Ph.D. degree in Photogrammetry and Remote Sensing from Wuhan University, Wuhan, China, in 2009. He was a Research Associate with the Department of Land Surveying and Geo-Informatics, The Hong Kong Polytechnic University, Hong Kong, in 2011, and a Visiting Scholar with the University of Illinois at Urbana-Champaign, Urbana, IL, USA, from 2013 to 2014. He is currently a Professor with the School of Geosciences and Info-Physics, Central South University, Changsha, China. He has authored over 30 journal articles. His current research interests include geo/remote sensing big data, machine/deep learning, and artificial/brain-inspired intelligence. 

%








\end{document}